\documentclass[11pt, a4paper, logo, copyright]{eai}
\usepackage{shlab}

\usepackage[numbers, sort&compress]{natbib}
\usepackage{array}
\usepackage{booktabs}
\usepackage{longtable}
\usepackage{amssymb}
\usepackage{amsmath}
\usepackage{tabularx}
\usepackage{multirow}
\usepackage{wrapfig}
\usepackage{ragged2e}
\usepackage{algorithm}
\usepackage{algpseudocode}
\usepackage{xspace}
\usepackage[most]{tcolorbox}
\usepackage{placeins}
\usepackage{cleveref}
\usepackage{titletoc}
\usepackage{tabularx}
\usepackage{makecell}
\usepackage{xcolor}
\usepackage{adjustbox}
\usepackage{subcaption}
\newcolumntype{Y}{>{\centering\arraybackslash}X}

\crefname{section}{Section}{Sections}
\Crefname{section}{Section}{Sections}
\crefname{subsection}{Section}{Sections}
\Crefname{subsection}{Section}{Sections}
\crefname{figure}{Fig.}{Figs.}
\Crefname{table}{Table}{Tables}
\crefname{table}{Tab.}{Tabs.}
\graphicspath{{}{appendix/}}

\newtcolorbox{promptbox}[1][]{
    enhanced,
    breakable,
    colback=blue!4,
    colframe=blue!25,
    boxrule=0.5pt,
    arc=2pt,
    left=6pt,
    right=6pt,
    top=5pt,
    bottom=5pt,
    fontupper=\small,
    #1
}

\newtcolorbox{negpromptbox}[1][]{
    enhanced,
    breakable,
    colback=red!4,
    colframe=red!25,
    boxrule=0.5pt,
    arc=2pt,
    left=6pt,
    right=6pt,
    top=5pt,
    bottom=5pt,
    fontupper=\small,
    #1
}

\usepackage{newfloat}
\usepackage{listings}
\DeclareCaptionStyle{ruled}{labelfont=normalfont,labelsep=colon,strut=off} 
\floatstyle{ruled}
\newfloat{listing}{tb}{lst}{}
\floatname{listing}{Listing}

\usepackage{booktabs}
\usepackage{amsmath}
\usepackage{amssymb}

\title{GAUGE: A Measurement-Grounded Benchmark for Physical Fidelity in Simulation Engines and Video World Models}

\newcommand{\corrauth}{\textsuperscript{\faEnvelope}}
\newcommand{\affmark}[1]{\textsuperscript{#1}}
\newcommand{\modelname}{\textsc{GAUGE}}
\newcommand{\tblYes}{\ensuremath{\checkmark}}
\newcommand{\tblPart}{\ensuremath{\triangle}}
\newcommand{\tblNo}{--}

\makeatletter
\renewcommand{\@author}{%
\begin{tabular}{c}
Shuai Wang\affmark{*,1}\quad Yaxin Feng\affmark{*,1,2}\quad Xuekun Jiang\affmark{*,1}\quad Shihan Tian\affmark{*,1}\quad Ningyu Yan\affmark{*,1,2}\\
Xing Shen\affmark{1}\quad Chaoyang Lyu\affmark{1}\quad Hui Wang\affmark{1,3}\quad Yunsong Zhou\affmark{1}\quad Hanqing Wang\affmark{1}\\
Jiangmiao Pang\affmark{1}\quad Yang Xiang\affmark{2}\quad Xing Gao\affmark{1}\corrauth\quad Chunhua Shen\affmark{1,4}\quad Weinan Zhang\affmark{1,3}
\end{tabular}\\[0.35em]
\makebox[\linewidth][c]{
\begin{tabular}{c}
\affmark{1}Shanghai Artificial Intelligence Laboratory\quad
\affmark{2}Hong Kong University of Science and Technology\\
\affmark{3}Shanghai Jiao Tong University\quad
\affmark{4}Zhejiang University
\end{tabular}}\\[0.25em]
}
\makeatother

\correspondingauthor={%
* Equal contribution.\\
\corrauth\ Corresponding author: Xing Gao (\texttt{gaoxing@pjlab.org.cn}).
}

\begin{document}

\begin{abstract}
Physics engines facilitate large-scale training and evaluation for embodied intelligence, while generative video world models are emerging as implicit simulators of future states and interactions. 
However, existing evaluations of physical fidelity are often conducted in isolation and rely heavily on perceptual similarity or human judgments, providing limited insight into which physical principles or parameters are violated.  
We introduce \modelname{}, a real-world-grounded diagnostic benchmark for jointly evaluating how numerical simulators and generative video world models reproduce or deviate from real-world physics. 
It comprises 22 controlled task families covering rigid bodies, flexible cables, textiles, and volumetric deformable objects. 
Grounded in real-world trajectories and paired with calibrated physical metadata, uncertainty annotations, and task-specific observables, these tasks cover  fundamental physical processes including collision, friction, momentum transfer, oscillation, self-contact, and deformation across diverse materials and conditions. 
We benchmark Isaac Sim, Genesis, and Newton on 14 task families using generalized trajectory errors, and evaluate 6 image-to-video models on 5 rigid-body tasks by testing physical-law consistency and the temporal stability of inferred parameters. 
Our results reveal no uniformly faithful physics engine, with the largest discrepancies arising in impulsive contact, rapid textile motion, and volumetric deformation. 
We further find that video world models can produce trajectories with the expected equation form while recovering incorrect accelerations, momentum transfer, and oscillation timing. 
\modelname{} lays the groundwork for developing more physically faithful simulators and world models for embodied intelligence.
Project page: https://internrobotics.github.io/GAUGE/.
\end{abstract}

\maketitle

\begin{figure*}[t]
\centering
\includegraphics[width=1\textwidth]{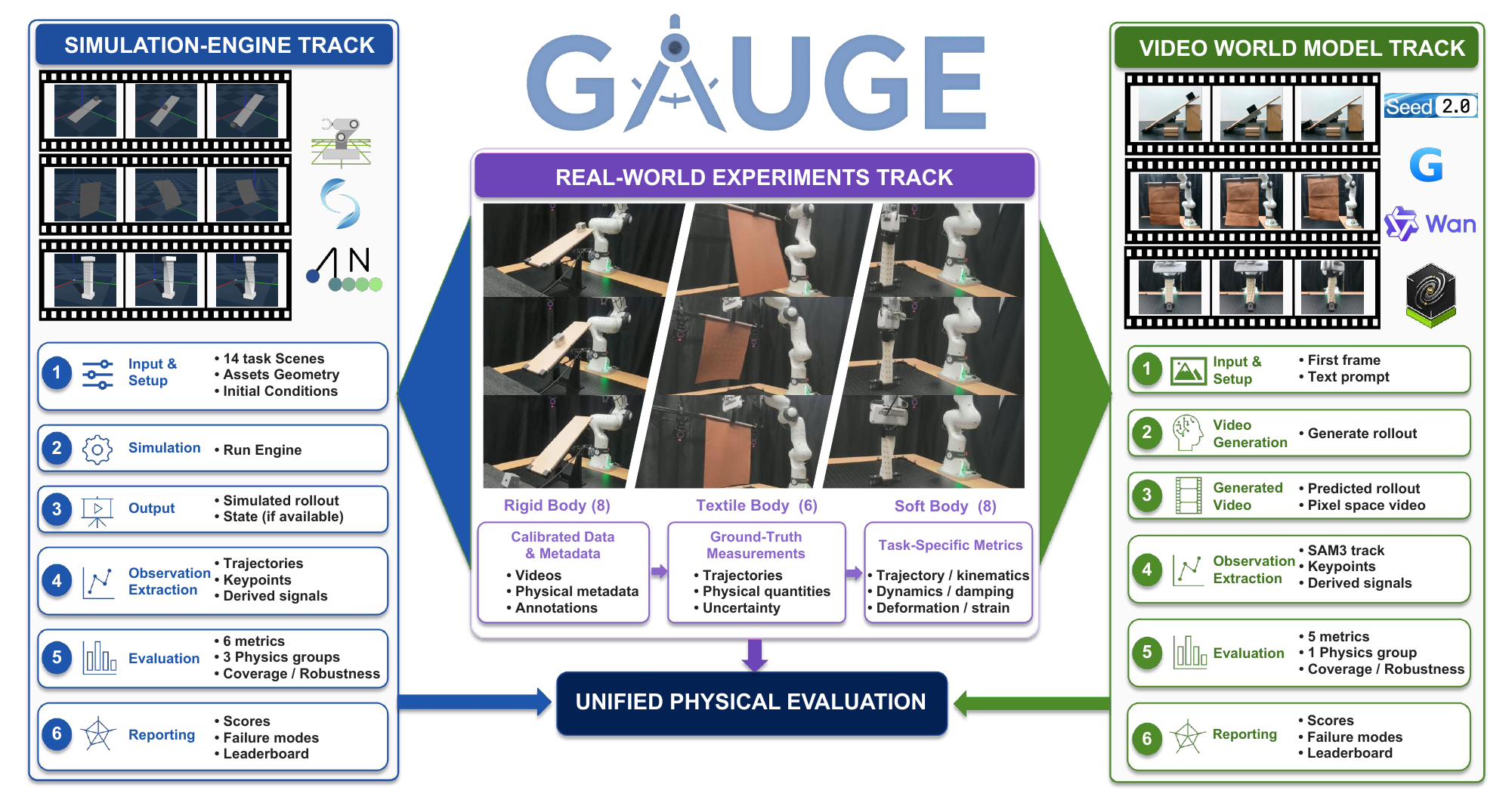} 
\caption{Overview of \modelname{}. Controlled real-world experiments provide calibrated physical metadata and motion-capture ground truth for rigid bodies, textiles, and volumetric deformable bodies. The physics-engine track (left) compares trajectories from matched simulated scenes with real measurements. The video world-model track (right) generates future motion from an initial frame and prompt, then evaluates the tracked motion in terms of physical laws and parameters.}
\label{fig1}
\end{figure*}


\section{Introduction}
Advances in embodied intelligence have benefited substantially from the widespread adoption of real-to-sim-to-real pipelines for robot learning, policy evaluation, and large-scale data generation.
Despite the approximations inherent in these pipelines, SIMPLER has demonstrated that carefully aligned simulation experiments can reliably predict the relative real-world performance of robot policies \cite{Torne2024, Li2024, Pfaff2025}. 
However, the effectiveness of such approaches depends critically on the physical fidelity of the underlying environment, whether it is instantiated by conventional numerical simulators, such as physics engines, or by generative approaches, such as video world models \cite{Wu2022, Bruce2024, Quevedo2025, Ye2026}. 
In particular, high visual fidelity does not necessarily imply accurate physical dynamics: a simulator may produce realistic-looking observations while misrepresenting motion, contact, or deformation, thereby encouraging policies to exploit simulation artifacts or yielding incorrect rankings of policy performance.

Unlike existing benchmarks that mainly measure visual realism or perceived physical plausibility, we introduce \modelname{}, a benchmark based on real-world experimental measurements for evaluating the physical fidelity of both physics engines and generative world models. 
As illustrated in \cref{fig1}, \modelname{} contains 22 standard experimental task families covering rigid bodies, one-dimensional flexible objects, quasi-two-dimensional textiles, and three-dimensional deformable bodies. 
These tasks evaluate collision, friction, momentum transfer, oscillation, self-contact, curvature, strain, and large deformation. 
For each task, millimeter-level observations, calibrated physical parameters, and estimates of measurement uncertainty are provided. 
To evaluate numerical simulators, we reconstruct several representative scenes in Isaac Sim, Genesis, and Newton, and compare their simulated results with real-world measurements. 
To evaluate generative world models, future videos are predicted by each model from initial scene images and textual prompts upon rigid-body settings. 
Through the trajectories recovered from video sequences,
GAUGE evaluates whether the generated motion conforms to the expected mathematical structure of the underlying physical law, enabling a more diagnostic assessment of both structural consistency and quantitative physical accuracy. 
Overall, \modelname{} serves as a unified diagnostic testbed for identifying not only whether a simulator or world model deviates from real-world physics, but also which physical mechanisms it fails to reproduce.

To summarize, our main contributions are listed below: 
\begin{itemize}
    \item \textbf{A unified real-world task suite.}
    We design 22 controlled task families spanning rigid bodies, flexible cables, textiles, and volumetric deformable objects. 
    The suite contains approximately 1,560 motion-capture trials, together with uncertainty estimates and task-specific physical observables.

    \item \textbf{Cross-regime physical parameter annotations.}
    To the best of our knowledge, \modelname{} is the first real-world benchmark dataset to provide task-associated, experimentally characterized parameters spanning rigid-body contact, cloth constitutive response, and volumetric soft-body mechanics within a single standardized collection.

    \item \textbf{Two complementary evaluation protocols.}
    Using the same real-world experimental foundation, we separately benchmark numerical physics engines and generative video world models. 
    The former measures task-specific sim-to-real discrepancies, while the latter distinguishes agreement with physical-law form from parameter accuracy and temporal stability.
\end{itemize}

\section{Related Work}
\paragraph{Real-World Validation of Physics Engines.}
Physical validation has progressed from rigid manipulation and impact studies to domain-specific deformable-object tests. Rigid-body work compares simulated and measured manipulation trajectories or high-speed impacts, showing that fidelity depends strongly on contact geometry, friction, and engine-specific calibration \cite{Collins2020, Acosta2021}. 
Other study focuses on calibration procedures rather than direct cross-engine validation \cite{Mehta2020}. 
For deformable systems, scaling-law experiments probe rods and plates under bending, twisting, buckling, and frictional contact \cite{Romero2021}; DiffCloud \cite{Sundaresan2022} uses real point clouds to identify simulator mass and stiffness scalings; and a cloth benchmark compares MuJoCo, Bullet, Flex, and SOFA under dynamic and quasi-static manipulation \cite{BlancoMulero2024}. 
PokeFlex \cite{Obrist2024} supplies synchronized meshes, point clouds, and interaction forces for volumetric objects , RGBench \cite{Hu2025} combines garment dynamics with instrument-measured fabric properties, and SORS \cite{Mekkattu2025} validates a soft-robot simulator through real experiments. 
Despite this progress, these efforts remain tied to individual object classes, and several fit effective simulator parameters from the same observations used for evaluation. 
Their results therefore do not support cross-regime comparison under a shared task and scoring definition.

\paragraph{Physical Consistency of Generative World Models.}
One line of work evaluates whether generated videos appear physically plausible. 
VideoPhy-2 \cite{Bansal2025} combines human judgments with learned evaluators, while PhyGenBench \cite{Meng2024}, WorldModelBench \cite{Li2025}, and PhyWorldBench \cite{Gu2025} use human or multimodal-model judgments over law- or scenario-specific criteria. 
These protocols scale to diverse phenomena, but their scores primarily indicate perceptual or commonsense agreement and do not directly localize errors in trajectories or physical parameters. 
A complementary line introduces reference-grounded tests. Controlled two-dimensional simulations expose weak out-of-distribution generalization of learned physical laws \cite{Kang2024}. 
Physics-IQ compares generated continuations with repeated real experiments, while Physics-IQ Verified audits its prompts, reference clips, and score aggregation \cite{Motamed2025, Raedsch2026}. 
PISA \cite{Li2025a} isolates free-fall behavior using mask- and trajectory-based metrics; WorldBench \cite{Upadhyay2026} combines real and synthetic tests of individual concepts and recovers gravity, friction, and viscosity from generated motion; and RIGIDBENCH \cite{jain2026} evaluates rigid-body motion against exact synthetic three-dimensional trajectories. 
These benchmarks make evaluation more quantitative, yet they typically trade broad coverage for coarse judgments, a narrow law set, or simulator-derived ground truth.

\paragraph{Physical Parameter Supervision.}
Physical metadata provides a complementary route to diagnosis, although existing annotations differ substantially in meaning. 
IRIS \cite{Khanbayov2026} pairs real videos with governing-equation labels and parameter targets for inverse recovery.
Although damping and friction values are fitted from trajectories, its collision coupling is an effective quantity rather than a measured contact parameter. 
PokeFlex \cite{Obrist2024} reports object-level effective stiffness rather than material elastic moduli, whereas RGBench \cite{Hu2025} measures cloth stretching and bending properties using dedicated instruments. 
FysicsEval \cite{Han2026} evaluates multimodal models on physical-attribute prediction and reasoning from real images, with instrument-based validation for selected properties, but does not assess generated dynamic rollouts. 
Thus, prior resources provide valuable physical quantities, but these quantities are generally confined to one mechanical regime, inferred from observed motion, or disconnected from dynamic-fidelity evaluation. 
In contrast, \modelname{} associates experimentally characterized rigid-contact, cloth-constitutive, and volumetric-soft-body parameters with repeated real state trajectories. 
To the best of our knowledge, it is the first standardized real-world benchmark to combine this cross-regime grounding with dedicated protocols for both physics engines and generative video world models. 
These parameters serve as task-matched metadata for scene reconstruction and failure diagnosis rather than as a separate parameter-estimation track.
\cref{tab1} provides a comprehensive comparison between \modelname{} and existing related benchmarks from multiple perspectives.

\begin{table*}[t]
\centering
\small
\setlength{\tabcolsep}{2.8pt}
\begin{tabularx}{\textwidth}{@{}>{\raggedright\arraybackslash}p{0.23\textwidth} *{9}{Y}@{}}
\toprule
&
\multicolumn{2}{c}{\textbf{Grounding}} &
\multicolumn{3}{c}{\textbf{Physical Regime}} &
\multicolumn{2}{c}{\textbf{Evaluated System}} &
\multicolumn{2}{c}{\textbf{Diagnosis}} \\
\cmidrule(lr){2-3}
\cmidrule(lr){4-6}
\cmidrule(lr){7-8}
\cmidrule(lr){9-10}

\textbf{Benchmark} &
\shortstack{\textbf{Real}\\\textbf{dyn.}} &
\shortstack{\textbf{Exp.}\\\textbf{param.}} &
\textbf{R} &
\textbf{C} &
\textbf{S} &
\textbf{Engine} &
\shortstack{\textbf{Video}\\\textbf{WM}} &
\shortstack{\textbf{Eq.}\\\textbf{form}} &
\shortstack{\textbf{Param.}\\\textbf{acc.}} \\
\midrule

Physics-IQ/Verified
\cite{Motamed2025, Raedsch2026}
& \tblYes & \tblNo
& \tblYes & \tblNo & \tblNo
& \tblNo & \tblYes
& \tblNo & \tblNo \\

IRIS \cite{Khanbayov2026}
& \tblYes & \tblPart
& \tblYes & \tblNo & \tblNo
& \tblNo & \tblNo
& \tblYes & \tblYes \\

WorldBench \cite{Upadhyay2026}
& \tblPart & \tblPart
& \tblYes & \tblNo & \tblNo
& \tblNo & \tblYes
& \tblPart & \tblYes \\

RIGIDBENCH \cite{jain2026}
& \tblNo & \tblNo
& \tblYes & \tblNo & \tblNo
& \tblNo & \tblYes
& \tblNo & \tblNo \\

Cloth Sim-to-Real \cite{BlancoMulero2024}
& \tblYes & \tblPart
& \tblNo & \tblYes & \tblNo
& \tblYes & \tblNo
& \tblNo & \tblNo \\

PokeFlex \cite{Obrist2024}
& \tblYes & \tblPart
& \tblNo & \tblNo & \tblYes
& \tblNo & \tblNo
& \tblNo & \tblNo \\

RGBench \cite{Hu2025}
& \tblYes & \tblYes
& \tblNo & \tblYes & \tblNo
& \tblYes & \tblNo
& \tblNo & \tblNo \\

FysicsEval \cite{Han2026}
& \tblNo & \tblPart
& \tblYes & \tblNo & \tblPart
& \tblNo & \tblNo
& \tblPart & \tblYes \\

\midrule
\textbf{\modelname{} (Ours)}
& \tblYes & \tblYes
& \tblYes & \tblYes & \tblYes
& \tblYes & \tblYes
& \tblYes & \tblYes \\
\bottomrule
\end{tabularx}
\caption{Comparison with representative datasets and benchmarks for physical grounding and physics-fidelity evaluation.
`\tblYes' denotes explicit support, `\tblPart' limited or indirect support, and `\tblNo' absence. ``Real dyn.'' denotes task-paired real dynamic
references. ``Exp. param.'' denotes task-associated experimental
characterization; fitted, effective, simulator-tuned, or restricted-subset
quantities are marked partial. R, C, and S denote rigid bodies, cloth,
and volumetric soft bodies, respectively. Fluids are omitted because
they are outside the current scope of \modelname{}.}
\label{tab1}
\end{table*}

\section{\modelname{}}
In this subsection, we present \modelname{}, our benchmark designed to provide a unified and standardized set of tasks for evaluating both numerical and generative simulators. 
The primary goal of \modelname{} is to enable fair and systematic assessment of a simulator's ability to model diverse physical properties under consistent evaluation settings. 
The design principles and data collection pipeline of the benchmark are described in detail in \cref{subsec:Dataset Pipeline}. 
Building upon this benchmark, we further evaluate both traditional physics engines and world models. 
As shown in \cref{fig1}, the corresponding pipelines are presented in \cref{subsec:Physics Engines Pipeline} and \cref{subsec:World Models Pipeline}.

\begin{table*}[!tp]
\centering
\small
\begin{tabular}{
  p{0.11\textwidth}
  p{0.27\textwidth}
  p{0.31\textwidth}
  p{0.07\textwidth}
  p{0.14\textwidth}
}
\toprule
\textbf{Task} &
\textbf{Description} &
\textbf{Evaluation} &
\textbf{Subtask} &
\textbf{Material} \\
\midrule

slope contact &
the fall of a trirectangular tetrahedron &
slope collisions &
1 &
wood, plastic \\
nonsmooth contact &
the fall of wedges and pyramids &
codimensional collisions &
3 &
wood, plastic \\
slope slider &
the slide of a cube on a slope &
static and kinetic friction &
3 &
wood, plastic, metal \\
turntable &
the slide of a cube on a turntable &
static and kinetic friction in a non-inertial frame.&
3 &
wood, plastic, metal \\
bouncing ball &
the fall and bouncing of a ball &
rapid impact &
1 &
rubber \\
newton's cradle &
the collision of closely fitted balls &
momentum transfer &
1 &
metal \\
wall breaking &
the collision of a wrecking ball with a wall &
large-scale dense collisions &
1 &
wood \\
pendulum &
the oscillation of a ball. &
periodic motion &
1 &
metal \\
rope winding &
the winding of a rope &
self-collision and stretch modulus &
1 &
rubber \\
textile stretching &
the tensile deformation of a textile &
stretch modulus &
1 &
R, S, U, O, L, N \\
textile bending &
the natural sagging of a textile &
bending modulus &
1 &
R, S, U, O, L, N \\
textile flinging &
the fluttering of a textile &
the motion under locally high acceleration &
1 &
R, S, U, O, L, N \\
funnel &
the passage of textiles through a hole &
collision, friction &
1 &
R, S, U, O, L, N \\
rotating ball &
the motion of a textile driven by a rotating ball &
static and kinetic friction &
1 &
R, S, U \\
tablecloth pulling &
the pulling of a textile from beneath rigid objects &
static and kinetic friction &
1 &
wood, plastic, metal \\
foam stretching &
the stretching of an elastic cuboid &
stretching modulus &
1 &
soft, hard \\
foam compressing &
the compression of an elastic cuboid &
compression modulus &
1 &
soft, hard \\
foam shearing &
the shearing of an elastic cuboid &
shear modulus &
1 &
soft, hard \\
foam twisting &
the twisting of an elastic cuboid &
twisting modulus &
1 &
soft, hard \\
foam bending &
the bending of an elastic cuboid &
bending modulus &
1 &
soft, hard \\
stick-stack &
the sliding of an elastic rod on a plane &
static and kinetic friction &
1 &
soft, hard \\
cantilever beam &
the free hanging of an elastic cantilever beam &
the coupling with a large stiffness ratio &
1 &
soft, hard \\
\bottomrule
\end{tabular}
\caption{The 22 \modelname{} task families, including each experiment's target physical property, number of condition variants, and material instantiations. The suite covers rigid bodies, rope, textiles, and volumetric deformable bodies. Abbreviations: R = rayon, S = satin, U = uniform cloth, O = Oxford fabric, L = synthetic leather, N = nylon taslan.}
\label{table1}
\end{table*}

\subsection{Dataset Pipeline}
\label{subsec:Dataset Pipeline}
\subsubsection{Dataset Design}
Unlike benchmarks designed for specific manipulation tasks, \modelname{} is constructed to evaluate simulators from a more fundamental physical perspective. 
To this end, we design 22 tasks that cover the distinct characteristics of rigid and deformable objects (\cref{app:subsec:standard scenes}).
Specifically, the benchmark includes 8 rigid-body tasks, which evaluate physical behaviors such as collision, friction, and the conservation of momentum and energy. 
For deformable objects, it contains 1 quasi-one-dimensional task (rope), 6 quasi-two-dimensional tasks (textile), and 7 three-dimensional tasks, covering key properties including stiffness, elastic modulus, rigid--deformable interaction, and friction. 
A complete list of tasks with their corresponding evaluation dimensions is provided in \cref{table1}.

To further increase the diversity of the benchmark, we introduce multiple material variants for each object category. 
Specifically, rigid objects are instantiated with three materials: wood, plastic, and metal. 
Textile objects include six commonly used fabric types: rayon, satin, uniform cloth, Oxford fabric, synthetic leather, and nylon taslan. 
For three-dimensional deformable objects, we consider both soft and hard variants of foam and rubber.

\subsubsection{Data Acquisition}
To accurately capture object states throughout the experiments, we use a millimeter-precision motion capture system to track object trajectories. 
For rigid objects, attached markers define a body-fixed frame at the object's geometric center, allowing us to recover its 6-DoF pose over time.
For deformable objects, supported by the skeletal tracking functionality, adjacent markers are connected to form a triangulated surface mesh, and their 3D positions are recorded throughout each trial, as shown in \cref{fig2}(b) and (c).

In addition to capturing the trajectories, \modelname{} provides detailed calibration of the physical properties of all assets involved in each task, including object dimensions, mass, and density. 
Specifically, we measure friction and restitution for rigid bodies using inclined-plane and collision tests. 
For textiles, tensile, shear, and bending stiffness are obtained from fabric tests conducted by Style3D.
And for three-dimensional deformable objects, Young's modulus and Poisson's ratio are calibrated using tensile tests and Digital Image Correlation (DIC) (see \cref{app:subsec:parameter calibration} for more details).
Based on the precise trajectory acquisition and comprehensive physical parameter calibration, we establish 22 standardized evaluation scenarios and evaluate the physical fidelity of both physics engines and video world models.

\subsection{Physics Engines Pipeline}
\label{subsec:Physics Engines Pipeline}
\subsubsection{Simulation and Track}
In \modelname{}, We evaluate Isaac Sim (v6.0.0), Genesis (v1.12.0), and Newton (v1.3.0) on 14 representative tasks. 
Each scene is reconstructed from the corresponding digital assets and motion-capture initial states before a rollout is generated. 
The solver depends on the object class. 
Rigid bodies use PhysX in Isaac Sim, the native solver in Genesis, and MuJoCo in Newton. 
Textiles use Surface FEM, PBD, and VBD, respectively, whereas volumetric deformable bodies use FEM, explicit and implicit MPM.
Because we aim to measure out-of-the-box performance, physical parameters are set from the \modelname{} calibrations, while all other settings remain at their defaults without task-specific fine-tuning.

During simulation data collection, the 6-DoF poses of rigid bodies can be directly exported from the simulation engine. 
However, Deformable tracking requires an additional step because simulated vertices or particles do not correspond to the motion-capture markers.
We therefore use the Hungarian algorithm to assign each initial marker to a unique nearby vertex or particle, with a matching error below $1~\mathrm{cm}$ \cite{Kuhn1955}.
The assigned elements are then tracked as marker proxies throughout the rollout.

\begin{figure}[!tp]
\centering
\includegraphics[width=0.5\textwidth]{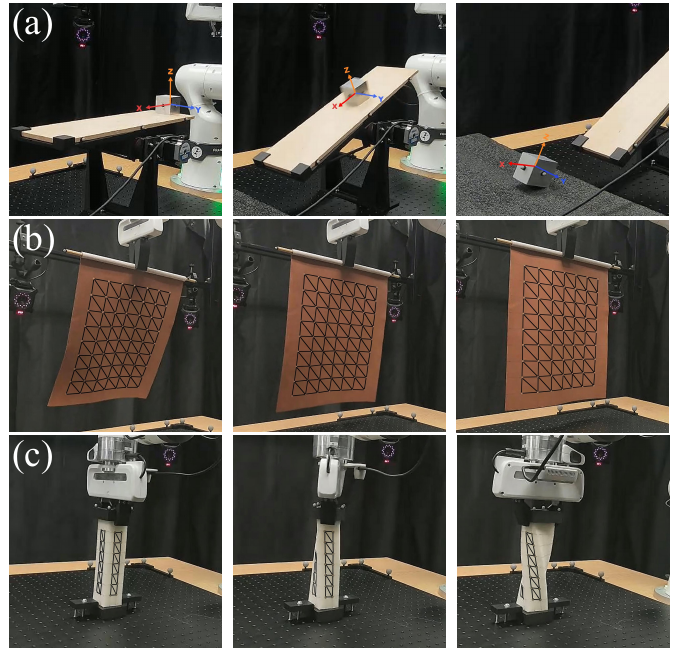} 
\caption{Generalized trajectories used for physics-engine evaluation. (a) Rigid-body motion is represented by object position $\mathbf{P}$; the example compares the real mean and uncertainty band with three simulated trajectories in the $X$--$Z$ plane. (b) Textile deformation is represented by marker-wise Gaussian curvature $\mathbf{K}$ on the tracked surface mesh. (c) Volumetric deformable body deformation is represented by the areas $\mathbf{A}$ of triangular faces formed by neighboring markers.}
\label{fig2}
\end{figure}

\subsubsection{Evaluation Metrics}
To compare different object classes within a common framework and preserve their physical features (details of the calculation method are deferred to \cref{app:subsec:metrics}), we represent each rollout by a generalized trajectory:
\begin{equation}
g(t)=
\begin{cases}
\mathbf{P}(t)\in\mathbb{R}^{3}, & \text{rigid bodies},\\
\mathbf{K}(t)\in\mathbb{R}^{N_m}, & \text{textiles},\\
\mathbf{A}(t)\in\mathbb{R}^{N_f}, & \text{deformable bodies}.
\end{cases}
\end{equation}
Here, $\mathbf{P}$ is the rigid-body position, while $\mathbf{K}$ and $\mathbf{A}$ collect the Gaussian curvatures at textile markers and the areas of deformable-body mesh faces. $N_m$ and $N_f$ denote the corresponding numbers of markers and faces. These representations are illustrated in \cref{fig2}.

Given a simulated trajectory $g_{\mathrm{sim}}$ and the mean trajectory $\bar{g}_{\mathrm{real}}$ over repeated real trials, we compute
\begin{equation}
\begin{aligned}
E_{\mathrm{RMSE}}
&=\sqrt{\frac{1}{T}\sum_{t=1}^{T}
\left\|g_{\mathrm{sim}}(t)-\bar{g}_{\mathrm{real}}(t)\right\|_2^2},\\
E_{\mathrm{DTW}}
&=\frac{1}{|\pi^\star|}
\sum_{(i,j)\in\pi^\star}
\left\|g_{\mathrm{sim}}(i)-\bar{g}_{\mathrm{real}}(j)\right\|_2 ,
\end{aligned}
\end{equation}
where $\pi^\star$ is the optimal dynamic time warping (DTW) alignment path. 
Root mean square error (RMSE) measures frame-aligned error, whereas DTW allows monotonic temporal alignment.
In \cref{table2}, the baseline reports the within-trial RMSE and per-trial RMSE standard deviation calculated from the real-world experimental data. 
Specifically, the within-trial RMSE is defined as $\mathrm{mean}_i[E_{\mathrm{RMSE}}(g_{\mathrm{real},i},\bar{g}_{\mathrm{real}})]$, which measures the typical deviation of an individual real trajectory from the mean trajectory. 
The per-trial RMSE standard deviation, defined as $\mathrm{std}_i[E_{\mathrm{RMSE}}(g_{\mathrm{real},i},\bar{g}_{\mathrm{real}})]$, characterizes the consistency of trajectory variations across different real-world trials.

For behaviors not fully captured by trajectory error, we introduce additional physical metrics for pendulum-related tasks. 
For Newton's cradle task:
\begin{itemize}
\item \textbf{Longest Stationary Duration (LSD)}: The maximum duration among all time intervals that satisfies the stationary condition, measuring whether the cradle exhibits the expected impulse-like motion.
\item \textbf{Momentum Transfer Efficiency (MTE)}: The ratio between the momentum of the outgoing ball and that of the incoming ball, evaluating whether momentum is fully transmitted during the collision process.
\end{itemize}
For the simple pendulum task:
\begin{itemize} 
\item \textbf{Period Duration (PD)}: The time required for the pendulum to complete one full oscillation, and measures whether the simulated oscillation period is consistent with the expected physical behavior. 
\item \textbf{Energy Loss (EL)}: The ratio of the final mechanical energy loss to the initial mechanical energy, evaluating whether the pendulum approximately conserves mechanical energy under the undamped setting.
\end{itemize}

\subsection{World Models Pipeline}
\label{subsec:World Models Pipeline}
\subsubsection{Generation and Track}
Because current world models still struggle with 3D consistency and deformable-object dynamics, we focus on rigid-body scenarios to evaluate six models: five image-to-video models---Cosmos3-Nano\cite{NVIDIA2026}, Cosmos3-Super-Image2Video (Cosmos3-Super-I2V)\cite{NVIDIA2026}, Wan-2.2\cite{Wan2025}, Wan-2.7, and Seedance 2.0\cite{Seedance2026}---and the interactive world model Genie 3. 
For each task, every model receives the same frontal initial frame and a standardized text prompt describing the scene, object properties, and expected motion, and then generates a future video.
It is worth noting that, for the Cosmos3 and Wan model families, we further conduct a controlled ablation by introducing a negative prompt shared across all models to constrain physically implausible behaviors and evaluate its impact on model inference. (\cref{app:subsubsec:standard input}).

To recover the motion from generated video sequences, we leverage SAM3\cite{Carion2025} to segment and track the target object. 
In general, we adopt text prompts to identify the target; when the segmentation is ambiguous, we add negative point prompts to exclude background regions. 
With reliable segmentation masks acquired, the object's 2D pixel coordinates are approximated from the centroid of each frame's binary segment mask.
Benefiting from its nearly constant depth relative to the camera, the object with known dimensions can provide an image-to-world scale. 
By utilizing this scale, we can thereby convert the centroid sequence into a two-dimensional trajectory in the experimental coordinate system (see \cref{app:subsubsec:SAM3} for more details).

\subsubsection{Evaluation Metrics}
Unlike previous evaluations of the physical consistency of world models, \modelname{} assesses whether a model has learned physical laws from two complementary perspectives: the structural form of the governing equations and their underlying physical parameters. 
This evaluation enables us to determine whether a model has truly captured the principles governing object motion, rather than merely generating visually plausible videos. 
To this end, we introduce the following metrics.

\begin{itemize}
\item \textbf{Dynamic Error (DE).}
Generally, for tasks with known external forces, DE measures consistency with Newton's second law:
\begin{equation}
\mathrm{DE}=\frac{1}{T}\sum_{t=1}^{T}
\left|m\ddot{x}_t-f_t\right|.
\end{equation}
Here, $T$ is the number of frames, while $m$, $x_t$, and $f_t$ denote object mass, position, and net external force at frame $t$. 
A lower DE indicates better force--motion consistency.
\item \textbf{Coefficient of Determination ($\mathbf{R^2}$).}
Whereas DE directly tests force--motion consistency, $\mathrm{R}^2$ measures how well the expected physical model explains the generated trajectory. 
A higher value indicates closer agreement with the expected equation form.
\item \textbf{Quadratic Form Improvement (QFI).}
QFI complements $\mathrm{R}^2$ by measuring the additional residual reduction obtained after adding a quadratic term to the expected linear relation. 
For uniformly accelerated motion (the free fall), displacement should be linear in squared time. 
Therefore, a low QFI indicates that the expected linear form is sufficient, whereas a high QFI reveals additional curvature.
The detailed computation of $R^2$ and QFI is provided in the \cref{app:subsec:physical analysis}
\end{itemize}

\section{Experiment Setup}
\subsection{Data Acquisition}
\begin{figure*}[h]
\centering
\includegraphics[width=0.8\textwidth]{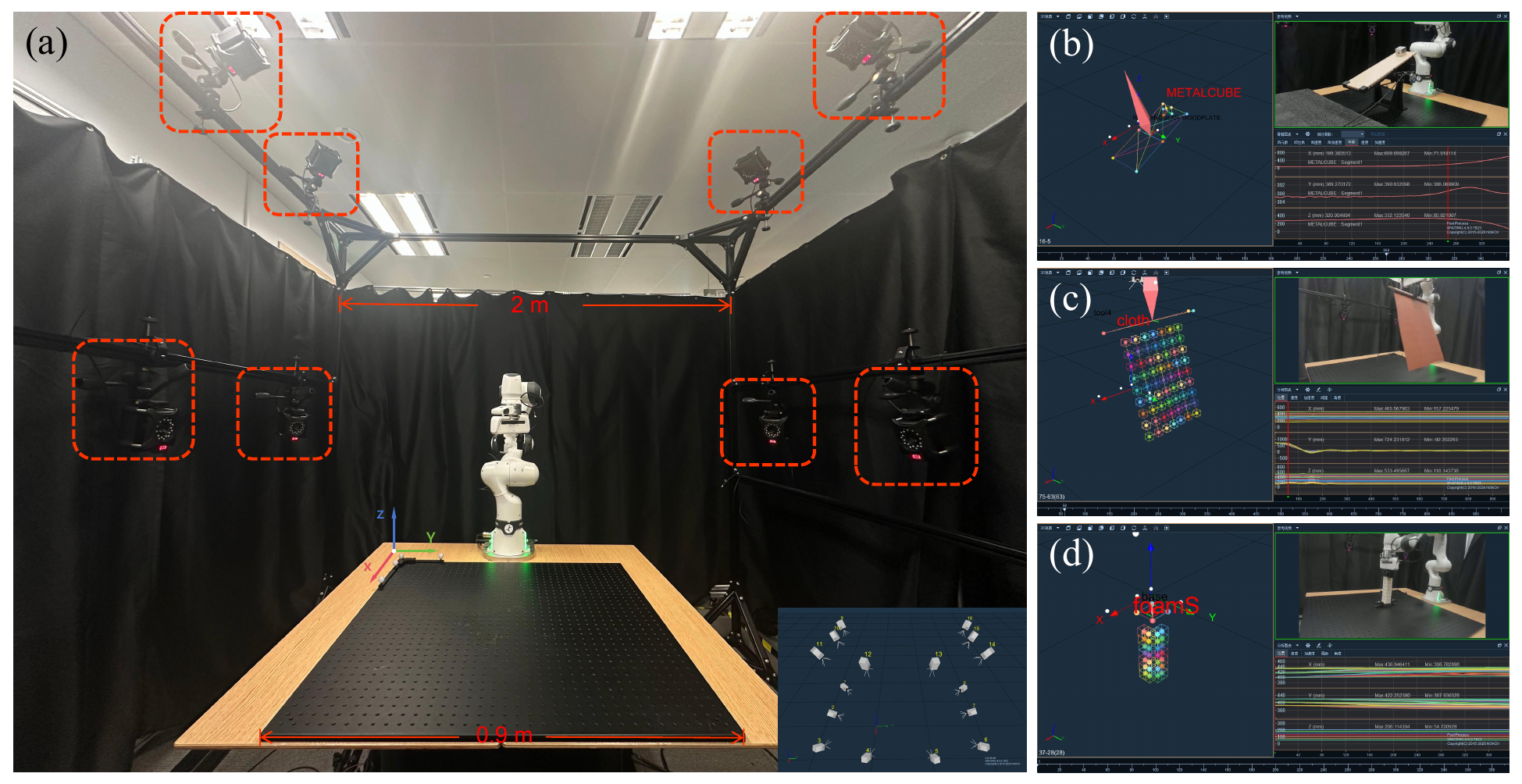} 
\caption{Real-world motion-capture system and representative tracking configurations.}
\label{fig3}
\end{figure*}

As illustrated in \cref{fig3}(a), we construct a $2~\mathrm{m}\times2~\mathrm{m}\times2~\mathrm{m}$ motion capture volume to accurately measure object trajectories in GAUGE. 
To fully cover the experimental workspace, we deploy 16 NOKOV Mars9H infrared cameras arranged across three height levels. 
The cameras operate at 180 Hz and provide sub-millimeter 3D localization accuracy. 
The experimental platform is built on a $0.9~\mathrm{m}\times0.9~\mathrm{m}$ black optical breadboard, where the world coordinate system is established using an L-shaped calibration fixture located at the upper-left corner of the workspace. 
To minimize environmental drift, all cameras are jointly calibrated with a T-shaped calibration fixture before each experiment.

Depending on the object category, different marker configurations and tracking strategies are adopted. 
As shown in \cref{fig3}(b), rigid-body objects are instrumented with $6~\mathrm{mm}$ retroreflective spheres (approximately $0.5~\mathrm{g}$ each), while reflective adhesive markers are used for spherical rigid objects. 
Since rigid bodies preserve their geometric structure, the motion capture system continuously estimates the pose of a rigid-body coordinate frame defined by multiple markers. 
In contrast, for textiles and deformable objects (\cref{fig3}(c) and (d)), $6~\mathrm{mm}$ reflective adhesive markers are attached to the object surface, and the system reconstructs local skeletal structures by connecting adjacent markers to track the trajectory of each marker individually. 
The masses of all markers are negligible and therefore have no measurable influence on the object dynamics.

After data acquisition, each recording is first truncated to the valid experimental interval and subsequently downsampled to 30 fps. 
For every evaluation task, we collect 20 independent real-world trials and store the processed trajectories together with the calibrated physical properties in JSON format for subsequent evaluation.

\subsection{Physics Engines}
During the physics-engine evaluation, we run all three engines on a Windows 11 workstation equipped with a single NVIDIA GeForce RTX 4090 GPU. 
The simulation frequency is set to 180 Hz for rigid-body and textile tasks, matching the motion capture frequency, and increased to 900 Hz for three-dimensional deformable-body tasks to prevent numerical divergence. 
Across all tasks, simulation states are recorded at 30 Hz.

\subsection{World Models}
During world-model inference, all models take the same initial image with a resolution of $832 \times 480$ as input (except for Genie3: $1300 \times 750$), while the generated videos retain each model's native output resolution and frame rate. 
Open-source models, including the Cosmos3 series and Wan-2.2, are deployed on the DLC distributed computing platform. 
Specifically, Cosmos3-Nano is inferred using a single NVIDIA GeForce RTX 4090 GPU, whereas Cosmos3-Super-I2V and Wan-2.2 each use eight NVIDIA GeForce RTX 4090 GPUs. 
Closed-source models are accessed through their official inference services: Wan-2.7 and Seedance 2.0 are invoked via remote APIs, while Genie 3 is generated directly through its online interface.

\section{Results}
\subsection{Physics Engines}

\begin{table*}[!h]
\centering
\small
\begin{tabular}{llrllrrr}
  \toprule
    \textbf{Task} & \textbf{Material} & \textbf{Frames} & \textbf{Metric} & \textbf{Baseline} & \textbf{IsaacSim} & \textbf{Genesis} & \textbf{Newton} \\
  \midrule
    slope contact & plastic & 10 & RMSE & 12.31 (7.62) & \textbf{1.26} & 1.60 & 1.75 \\
     &  &  & DTW & 5.30 (1.84) & \textbf{1.83} & 2.57 & 3.07 \\
  \midrule
    nonsmooth contact & plastic & 20 & RMSE & 17.67 (6.19) & \textbf{1.53} & 10.04 & 6.05 \\
     &  &  & DTW & 8.92 (2.44) & \textbf{1.90} & 14.37 & 8.01 \\
  \midrule
    slope slider & metal & 60 & RMSE & 38.01 (28.53) & 0.61 & \textbf{0.58} & 1.95 \\
     &  &  & DTW & 9.54 (8.52) & 0.71 & \textbf{0.69} & 2.93 \\
  \midrule
    turntable & metal & 95 & RMSE & 13.26 (7.43) & \textbf{0.17} & 0.57 & 20.04 \\
     &  &  & DTW & 3.33 (0.81) & \textbf{0.61} & 2.01 & 66.72 \\
  \midrule
    bouncing ball & rubber & 18 & RMSE & 5.29 (3.60) & \textbf{15.63} & 22.50 & 22.71 \\
     &  &  & DTW & 4.23 (2.87) & \textbf{5.58} & 14.58 & 14.89 \\
  \midrule
    newton's cradle & metal & 1027 & LSD & 0.38 (0.015) & 0.00 & -- & 0.00 \\
     &  &  & MTE & 93.02 (2.47) & 0.20 & -- & \textbf{0.26} \\
  \midrule
    pendulum & metal & 300 & PD & 1.14 (0.066) & 1.10 & 2.47 & \textbf{1.09} \\
     &  & 3000 & EL & 0.00 & 0.034 & \textbf{$-$0.041} & 0.022 \\
  \midrule
    textile stretching & satin & 380 & RMSE & 0.21 (0.42) & \textbf{0.73} & 1.51 & \textbf{0.73} \\
     &  &  & DTW & 0.21 (0.42) & 0.76 & 1.56 & \textbf{0.75} \\
  \midrule
    textile bending & satin & 335 & RMSE & 1.92 (0.69) & \textbf{7.94} & 11.73 & 19.90 \\
     &  &  & DTW & 1.20 (0.17) & \textbf{11.78} & 17.35 & 18.82 \\
  \midrule
    textile flinging & satin & 562 & RMSE & 0.016 (0.0056) & 128.26 & \textbf{8.54} & 9.25 \\
     &  &  & DTW & 0.012 (0.0014) & 167.31 & \textbf{11.27} & 12.12 \\
  \midrule
    foam stretching & soft & 30 & RMSE & 8.72 (6.94) & 16.15 & \textbf{10.05} & 15.46 \\
     &  &  & DTW & 8.19 (6.93) & 17.18 & \textbf{10.65} & 16.45 \\
  \midrule
    foam shearing & soft & 37 & RMSE & 5.60 (0.70) & 26.13 & \textbf{15.26} & 26.57 \\
     &  &  & DTW & 5.07 (0.59) & 28.85 & \textbf{16.74} & 29.34 \\
  \midrule
    foam twisting & soft & 65 & RMSE & 8.07 (0.53) & 17.14 & \textbf{10.86} & 16.78 \\
     &  &  & DTW & 7.48 (0.39) & 18.47 & \textbf{11.67} & 18.07 \\
  \midrule
    foam bending & soft & 35 & RMSE & 7.00 (4.39) & 10.71 & 14.43 & \textbf{10.37} \\
     &  &  & DTW & 3.52 (1.05) & 21.28 & 27.51 & \textbf{20.53} \\
  \bottomrule
\end{tabular}
\caption{Physical fidelity of Isaac Sim, Genesis, and Newton on 14 tasks. Baseline reports the real-world mean and standard deviation. Simulator entries are normalized by the baseline mean: lower is better for RMSE and DTW, while values closer to one are better for normalized LSD, MTE, and PD. Bold marks the best valid result, and a dash denotes an unavailable rollout. The pendulum EL row reports raw values because its baseline is zero; zero is ideal.}
\label{table2}
\end{table*}

\paragraph{Rigid-body dynamics.}
\cref{table2} reports each simulator's discrepancy relative to the mean value of the baseline.
For RMSE and DTW, a normalized value below one indicates that the simulator-to-real discrepancy is smaller than the within-trial RMSE measured across repeated real-world trials.
Isaac Sim gives the lowest errors for slope contact, nonsmooth contact, and turntable motion.
Its turntable errors are particularly small, with normalized RMSE and DTW of 0.17 and 0.61, respectively.
Genesis performs best on the slope-slider task, reaching 0.58 RMSE and 0.69 DTW, while Newton is less accurate in this friction-dominated setting.
These results show that standard contact and sliding behaviors can be reproduced reasonably well, but the ranking depends on the contact geometry and reference frame.

More demanding rigid-body events expose a substantially larger gap.
Even the best bouncing-ball result is 15.63 times the baseline RMSE and 5.58 times the baseline DTW.
For Newton's cradle, Isaac Sim and Newton produce zero longest stationary duration, and their normalized momentum-transfer efficiencies are only 0.20 and 0.26; Genesis does not produce a valid rollout.
The pendulum is easier in terms of frequency: Isaac Sim and Newton obtain normalized periods of 1.10 and 1.09.
However, all engines accumulate energy error over the longer rollout, with raw energy-loss values ranging from $-0.041$ to $0.034$.
Thus, matching low-frequency motion does not imply accurate impact resolution or long-horizon energy behavior.

\paragraph{Textile and deformable body dynamics.}
The textile results vary markedly with the deformation regime.
For textile stretching, Isaac Sim and Newton achieve normalized errors close to or below one.
For textile bending, however, the best RMSE and DTW increase to 7.94 and 11.78.
The largest solver-specific failure occurs in textile flinging: Genesis obtains the lowest RMSE of 8.54, whereas Isaac Sim reaches 128.26.
This contrast indicates that quasi-static agreement does not reliably predict fidelity under rapid, spatially varying deformation.
For volumetric deformable bodies, Genesis gives the lowest errors for stretching, shearing, and twisting, while Newton performs best for bending.
Nevertheless, even the best-performing deformable body simulations exhibit errors approximately one order of magnitude above the corresponding real-world baselines.
Overall, no engine dominates across rigid, textile, and volumetric regimes.
The results instead reveal complementary solver strengths and identify dynamic contact, high-acceleration cloth motion, and three-dimensional deformation as the main sources of residual sim-to-real error.

\subsection{World Models}

\begin{table*}[!tp]
\centering
\small
\setlength{\tabcolsep}{1mm}
\resizebox{\linewidth}{!}{%
\begin{tabular}{lllrrrrrrrrrrr}
\toprule
\textbf{Task} &
\textbf{Mat.} &
\textbf{Met.} &
\textbf{Base.} &
\multicolumn{2}{c}{\textbf{Cosmos3-N}} &
\multicolumn{2}{c}{\textbf{Cosmos3-S}} &
\multicolumn{2}{c}{\textbf{Wan22}} &
\multicolumn{2}{c}{\textbf{Wan27}} &
\textbf{Seed2} &
\textbf{Genie3} \\
\cmidrule(lr){5-6}
\cmidrule(lr){7-8}
\cmidrule(lr){9-10}
\cmidrule(lr){11-12}
& & & &
\textbf{w/o} & \textbf{w} &
\textbf{w/o} & \textbf{w} &
\textbf{w/o} & \textbf{w} &
\textbf{w/o} & \textbf{w} &
\textbf{w/o} & \textbf{w/o} \\
\midrule

slope slider
& wood
& QFI
& 0.00
& 64.89
& 26.65
& \textbf{13.61}
& 569.36
& 393.00
& 15.74
& 1201.08
& 558.97
& 781.51
& 40.11 \\

&
& $a$
& 2.58
& \textbf{2.06}
& 0.90
& 0.13
& 0.12
& 0.43
& 0.091
& 0.061
& 0.038
& 0.23
& 0.62 \\

& plastic
& QFI
& 0.00
& 129.76
& 276.49
& 25.60
& 330.31
& 14.24
& 40.12
& 1164.59
& 942.08
& \textbf{8.69}
& 67.07 \\

&
& $a$
& 2.57
& 0.58
& 0.22
& 0.36
& 0.030
& \textbf{0.75}
& 0.16
& 0.055
& 0.063
& 0.20
& 0.58 \\

& metal
& QFI
& 0.00
& \textbf{4.41}
& 184.09
& 1266.40
& 924.39
& 116.64
& 6.96
& 60.70
& 1145.94
& 730.22
& 70.37 \\

&
& $a$
& 2.67
& 0.36
& 0.15
& 0.056
& 0.068
& 0.24
& 0.40
& 0.046
& 0.038
& 0.074
& \textbf{0.43} \\

\midrule

turntable
& wood
& DE
& 0.00
& 0.078
& 0.10
& 0.092
& \textbf{0.00}
& \textbf{0.00}
& \textbf{0.00}
& 0.18
& 0.00037
& \textbf{0.00}
& \textbf{0.00} \\

& plastic
& DE
& 0.00
& 0.11
& 0.17
& 0.0029
& 0.11
& \textbf{0.00}
& \textbf{0.00}
& 0.032
& 0.050
& \textbf{0.00}
& \textbf{0.00} \\

& metal
& DE
& 0.00
& 0.0064
& 0.19
& 0.040
& 0.25
& 0.00048
& \textbf{0.00}
& 0.12
& 0.78
& 0.0024
& \textbf{0.00} \\

\midrule

bouncing ball
& rubber
& QFI
& 0.00
& 2620.05
& 1125.36
& 270.69
& \textbf{12.50}
& 428.39
& 38.72
& 59.49
& 16.08
& 65.16
& 88.70 \\

&
& $a$
& 9.81
& 0.45
& 0.33
& 0.076
& 0.088
& 0.051
& 0.064
& 0.16
& 0.15
& \textbf{1.84}
& $-$0.052 \\

\midrule

newton's cradle
& metal
& MTE
& $\sim$1
& ---
& ---
& ---
& ---
& ---
& \textbf{0.76}
& 0.55
& 0.48
& 0.24
& --- \\

\midrule

pendulum
& metal
& $\mathrm{R}^2$
& 1.00
& 0.86
& 0.66
& 0.50
& 0.92
& 0.98
& \textbf{0.99}
& 0.71
& 0.72
& 0.96
& \textbf{0.99} \\

&
& $\mathrm{PD}$
& 1.06
& 2.36
& 2.93
& 17.95
& 2.34
& 6.03
& 1.93
& \textbf{1.83}
& 1.87
& 3.13
& 1.90 \\
\bottomrule
\end{tabular}%
}
\caption{Physical consistency of video world models on five rigid-body tasks. \textbf{w/o} and \textbf{w} denote generation without and with the negative prompt. Lower is better for QFI and DE; acceleration $a$, MTE, and PD are compared with the real baseline; and higher is better for $\mathrm{R}^2$. Bold marks the best valid result, and dashes denote unavailable generations. Abbreviations: Mat.\ = material, Met.\ = metric, Base.\ = baseline, Cosmos3-N/S = Cosmos3-Nano/Super-I2V, Seed2 = Seedance 2, $a$ in $\mathrm{m/s^2}$, and $\mathrm{PD}$ in seconds.}
\label{table3}
\end{table*}

\begin{figure*}[t]
\centering
\begin{minipage}[t]{0.32\textwidth}
  \centering
  \includegraphics[width=\linewidth]{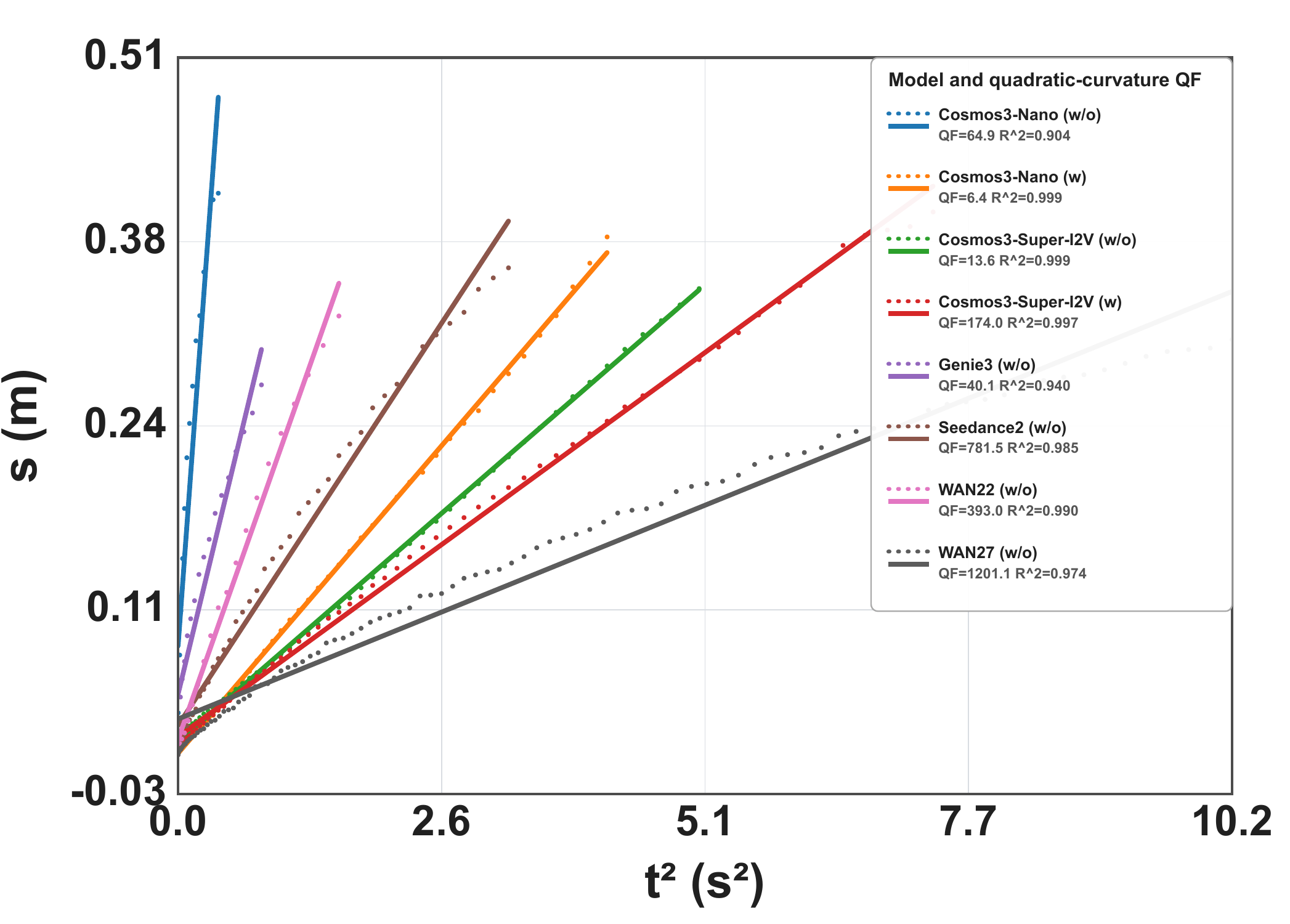}\\
  \small (a) Slope slider
\end{minipage}
\hfill
\begin{minipage}[t]{0.32\textwidth}
  \centering
  \includegraphics[width=\linewidth]{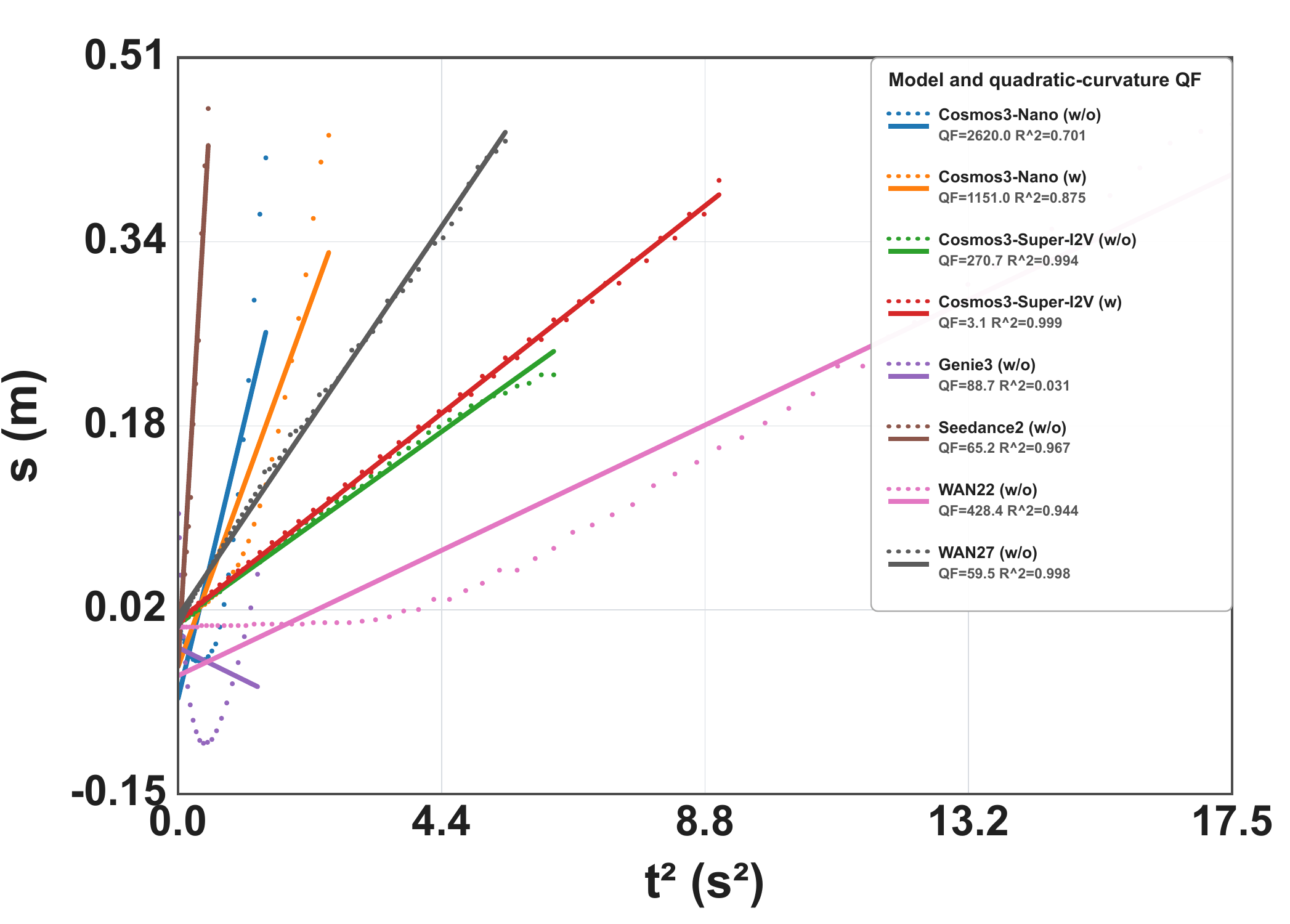}\\
  \small (b) Free fall
\end{minipage}
\hfill
\begin{minipage}[t]{0.32\textwidth}
  \centering
  \includegraphics[width=\linewidth]{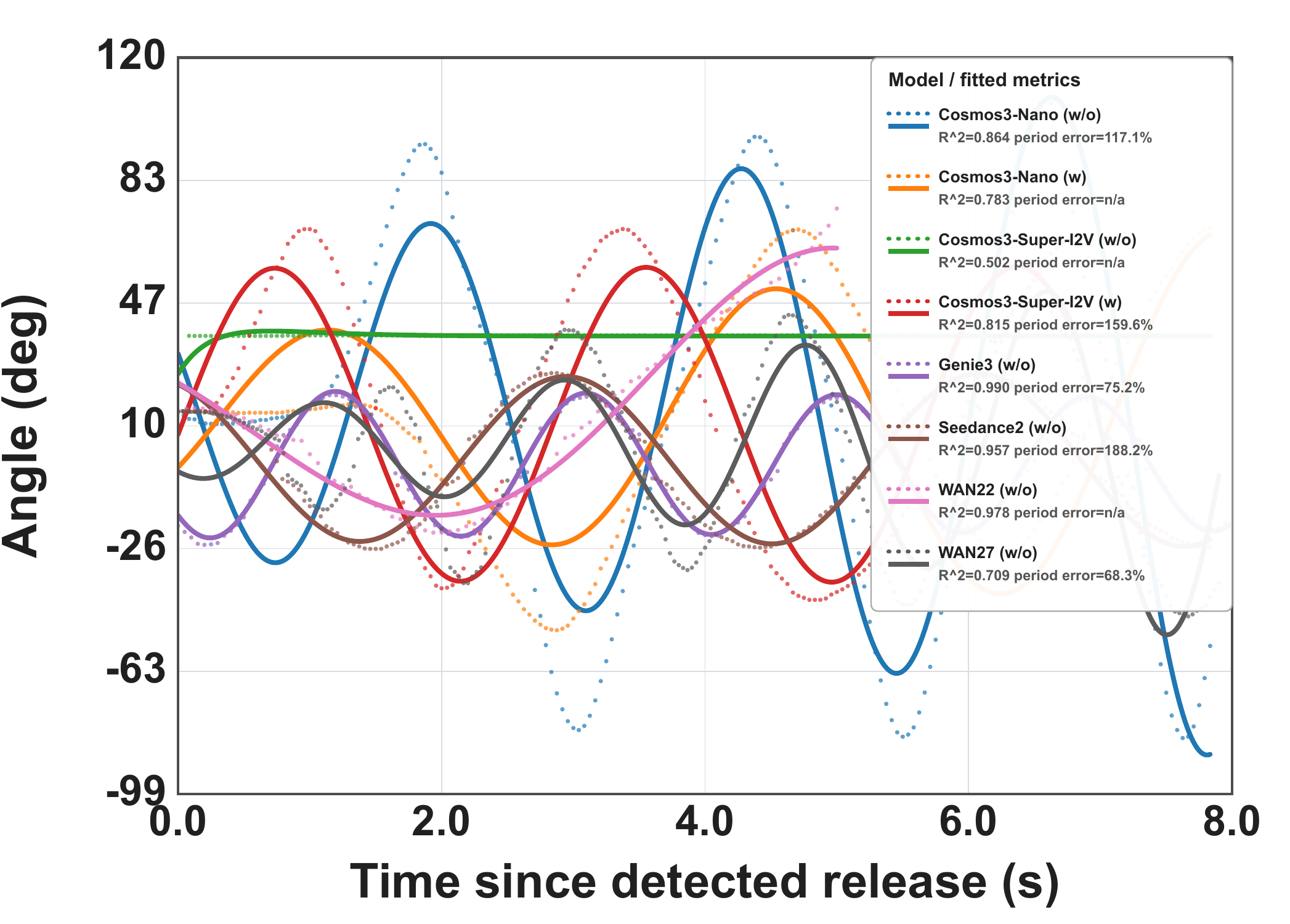}\\
  \small (c) Pendulum
\end{minipage}
\caption{Representative physical-law fits for generated videos. Displacement is plotted against squared time for (a) the wood slope slider and (b) the free-fall segment of the bouncing ball; uniform acceleration should produce a straight line. (c) Tracked pendulum angles and fitted oscillations after release. Dotted curves are recovered trajectories and solid curves are physical-model fits; the legends report QFI or fitted parameters. Differences in slope, curvature, period, and damping reveal errors not captured by visual plausibility alone.}
\label{fig4}
\end{figure*}

\paragraph{Law form and parameter accuracy.}
\cref{table3} and \cref{fig4} shows that satisfying the structural form of a physical law and recovering its parameters are distinct capabilities.
For slope sliding, the lowest QFI is achieved by different models for wood, plastic, and metal: Cosmos3-Super-I2V, Seedance 2, and Cosmos3-Nano, respectively.
This change in ranking indicates limited consistency across materials.
More importantly, a low QFI does not guarantee a correct acceleration.
For wood, the best estimate, $2.06\,\mathrm{m/s^2}$, is reasonably close to the measured $2.58\,\mathrm{m/s^2}$.
For plastic and metal, however, the closest reported accelerations are only 0.75 and $0.43\,\mathrm{m/s^2}$, compared with real values of 2.57 and $2.67\,\mathrm{m/s^2}$.
\cref{fig4} (a) visualizes this discrepancy: several trajectories are approximately linear in squared time, yet their slopes differ substantially.

The bouncing-ball task provides an even clearer example.
Cosmos3-Super-I2V with the negative prompt obtains the lowest QFI of 12.50, but its inferred acceleration is only $0.088\,\mathrm{m/s^2}$.
Seedance 2 gives the closest acceleration among the evaluated models, but its value of $1.84\,\mathrm{m/s^2}$ remains far below gravitational acceleration.
As shown in \cref{fig4} (b), the generated motion can resemble a uniformly accelerated trajectory over a short interval while encoding an incorrect physical scale.
This result supports evaluating equation structure and parameter accuracy separately.

\paragraph{Interaction and oscillation.}
Momentum transfer remains difficult.
Six of the ten reported model configurations fail to produce a valid Newton's-cradle sequence, and Wan-2.2 with the negative prompt reaches the highest MTE of 0.76, below the real-world value of approximately one.
The pendulum results further illustrate the limits of perceptual plausibility.
Wan-2.2 and Genie3 attain an $\mathrm{R}^2$ of 0.99, indicating a strong oscillatory fit, but their periods are 1.93~s and 1.90~s rather than the measured 1.06~s.
The closest period is 1.83~s from Wan-2.7, which is still about 73\% longer than the baseline.
\cref{fig4} (c) further shows substantial variation in fitted damping and amplitude, including different damping signs across models.
Consequently, a generated sequence may look periodic and achieve a high fitting score while still violating the underlying temporal parameters and energy behavior.

\paragraph{Prompt sensitivity.}
The paired settings in \cref{table3} show that the negative prompt does not yield a consistent improvement.
For example, it reduces the bouncing-ball QFI of Cosmos3-Super-I2V from 270.69 to 12.50 and enables Wan-2.2 to produce a Newton's-cradle rollout, but it increases the wood slope-slider QFI of Cosmos3-Super-I2V from 13.61 to 569.36.
The direction and magnitude of the change depend on both the model and the task.
This sensitivity suggests that physics evaluation should use fixed prompt templates and report paired results rather than drawing conclusions from a single favorable generation.
More analysis details can be found in \cref{app:subsec:physical analysis}.

\section{Limitations and Outlook}
\modelname{} has several limitations that also define directions for future work.
First, although the current benchmark covers rigid bodies, textiles, and three-dimensional deformable bodies, its materials and calibrated parameter ranges remain limited.
Materials within the same nominal category can differ because of surface treatment, internal structure, manufacturing process, temperature, and wear.
Future versions should therefore include more materials and broader ranges of friction, stiffness, density, restitution, and damping.
The task set should also be extended to fluids and coupled processes, including fluid-rigid and fluid-soft-body interactions, to better represent the physical diversity encountered by embodied agents.

Second, the current world-model track focuses on rigid-body tasks that can be evaluated from two-dimensional image trajectories.
This representation is insufficient for textiles and volumetric deformable bodies, whose states involve distributed deformation, depth variation, self-occlusion, and self-contact.
When reliable three-dimensional coordinates become available, future metrics should operate on reconstructed point clouds, meshes, or tracked surface elements.
Relevant measurements include local strain, area and volume change, Gaussian and mean curvature, bending energy, geodesic distortion, self-contact, oscillation modes, damping, and energy dissipation.
Correspondence and reconstruction uncertainty must also be modeled explicitly so that errors in three-dimensional perception are not incorrectly attributed to the world model.
Combining these measurements with \modelname{}'s law-form and parameter-stability analysis would enable a more accurate evaluation of generated textile and deformable body dynamics.

\section{Conclusion}
We presented \modelname{}, a measurement-grounded benchmark for evaluating the physical fidelity of both physics engines and generative video world models.
\modelname{} contains 22 controlled task families spanning rigid bodies, flexible cables, textiles, and volumetric deformable objects, together with repeated real-world trajectories, calibrated physical parameters, and task-specific observables.
The benchmark supports a common evaluation framework while retaining metrics that are meaningful for different physical regimes.

Our experiments show that current physics engines have strongly domain-dependent accuracy.
Isaac Sim performs well on several rigid contact tasks, Genesis is more competitive on dynamic textile and most deformable body tasks, and Newton is strongest in selected deformation cases; none is consistently accurate across all regimes.
Current video world models exhibit a related but different limitation.
They can generate trajectories that are visually plausible or well fitted by a physical equation while predicting incorrect accelerations, momentum-transfer efficiencies, oscillation periods, or damping behavior.
Their performance is also sensitive to prompt formulation.
These findings demonstrate that physical fidelity cannot be characterized by visual quality, a single trajectory distance, or parameter estimation alone.
By jointly evaluating law form, physical parameters, and temporal behavior against real measurements, \modelname{} provides a diagnostic basis for improving simulators and world models intended for embodied intelligence.

\bibliography{gauge}

\clearpage
\appendix
\renewcommand{\thesection}{\Alph{section}}
\renewcommand{\thesubsection}{\thesection.\arabic{subsection}}
\startcontents[appendix]
\section*{Appendix Contents}
\printcontents[appendix]{}{1}{\setcounter{tocdepth}{3}}

\clearpage
\section{Benchmark dataset}
\subsection{Standard Scenes}
\label{app:subsec:standard scenes}
\begin{figure*}[!h]
\centering
\includegraphics[width=0.8\textwidth]{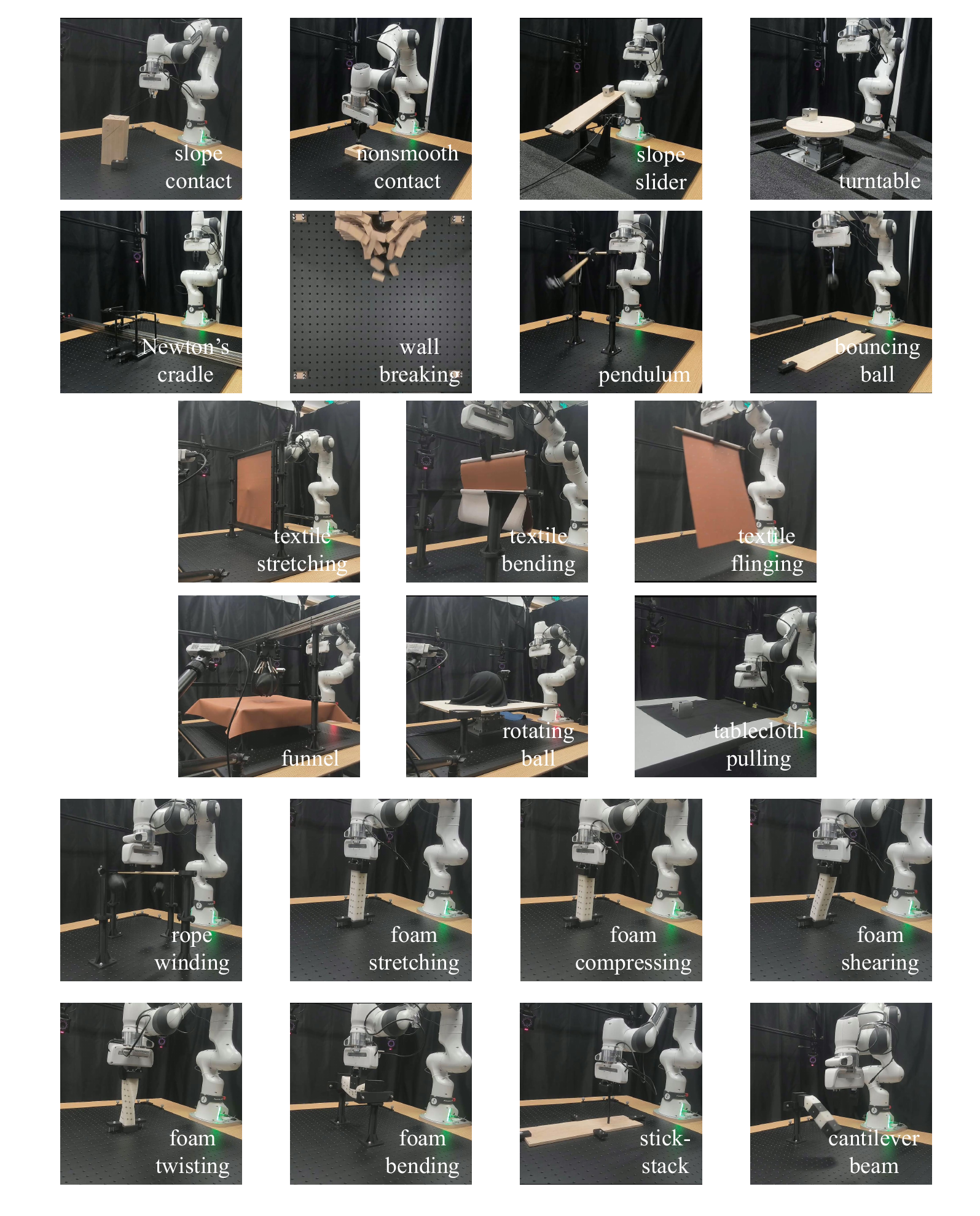} 
\caption{Overview of \modelname{}'s 22 standardized real-world task families.}
\label{s1}
\end{figure*}
\cref{s1} provides an overview of all 22 standardized scenarios included in \modelname{}, consisting of 8 rigid-body tasks, 6 textile tasks, and 8 deformable-body tasks. 
For rigid-body scenarios, we focus on evaluating fundamental physical interactions, including collision, friction, elasticity, and momentum/energy transfer. 
For textile and deformable-body scenarios, we primarily investigate self-collision, various stiffness-related properties, and rigid--deformable coupling behaviors.

\begin{figure*}[!h]
\centering
\includegraphics[width=0.5\textwidth]{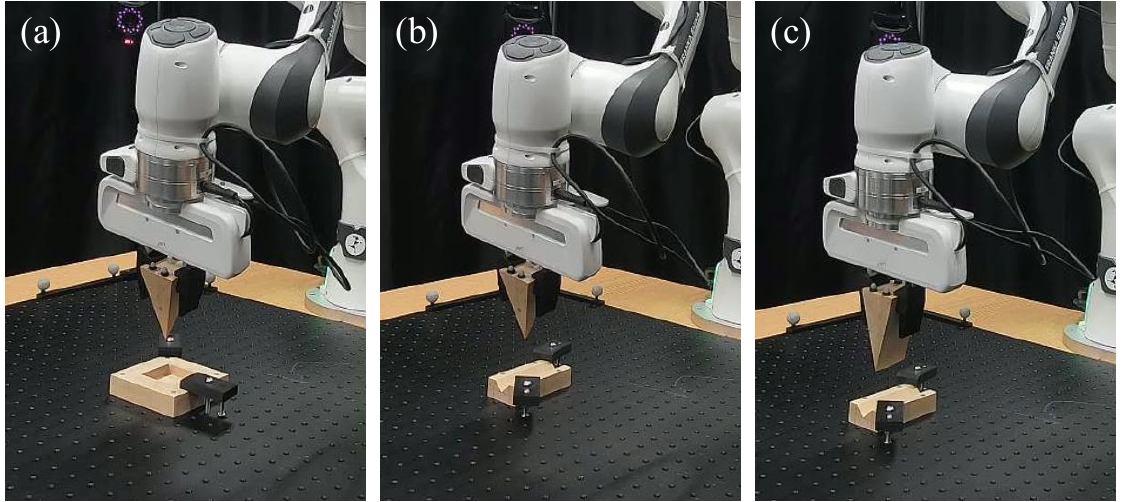} 
\caption{Geometry variants of the nonsmooth contact task.}
\label{s2}
\end{figure*}

Notably, three tasks, namely "Nonsmooth Contact", "Slope Slider", and "Turntable", each contain three sub-tasks. 
The three sub-tasks of "Nonsmooth Contact" correspond to collisions between (1) a quadrangular pyramid and a square-groove base, (2) a quadrangular pyramid and a triangular-groove base, and (3) a wedge and a triangular-groove base, respectively (\cref{s2}). 
The three sub-tasks of "Slope Slider" and "Turntable" are designed with different slope inclinations and rotational velocities, respectively, causing the cube to undergo different friction regimes, including static friction, dynamic friction, and transitions between them.

\subsection{Parameter calibration}
\label{app:subsec:parameter calibration}
\subsubsection{Parameter Measurements}
Beyond providing a comprehensive set of scenarios covering rigid bodies, textiles, and deformable objects, GAUGE further distinguishes itself by providing detailed physical property calibration for each scenario. 
In addition to basic information such as mass and geometric dimensions, GAUGE measures critical physical parameters, including friction coefficients and restitution coefficients for rigid bodies, stretch and bending stiffness for textiles, and Young's modulus and Poisson's ratio for deformable materials.

\begin{figure*}[!tp]
\centering
\includegraphics[width=0.8\textwidth]{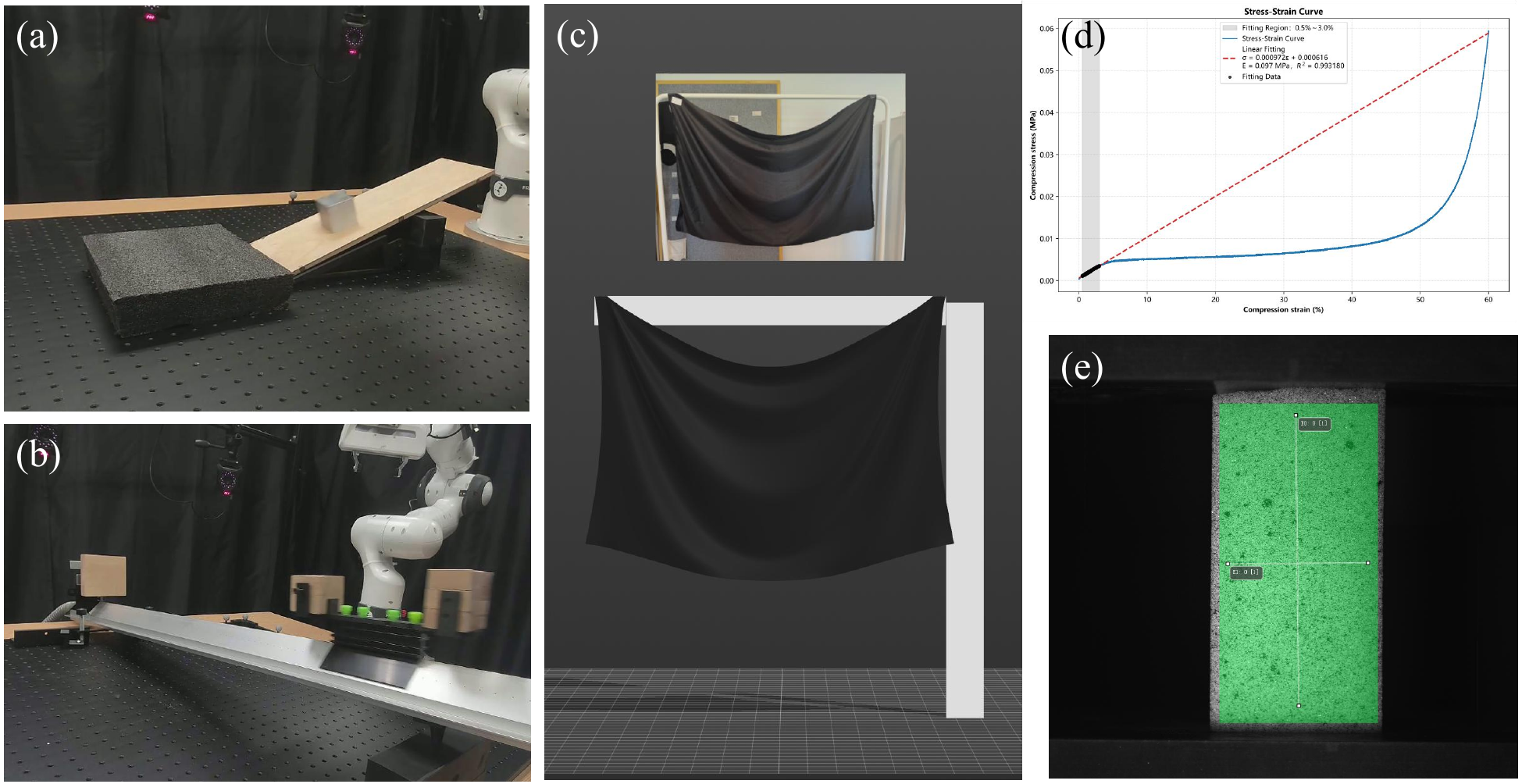} 
\caption{Experimental characterization and validation of task-specific physical parameters.}
\label{s3}
\end{figure*}

For friction coefficient measurement between rigid bodies (as well as between rigid bodies and textiles), we mainly adopt the inclined plane method (\cref{s3}(a)).
By measuring the sliding velocity $v(t)$ of an object, the friction coefficient is calculated as: $\mu=(g\sin\theta-\dot{v})/(g\cos\theta)$, where $g$ denotes the gravitational acceleration and $\theta$ represents the inclination angle of the slope. 
Furthermore, we measure the coefficient of restitution using a collision-based method. 
To eliminate friction and ensure consistent initial collision velocities, we employ an inclined air-bearing rail as illustrated in \cref{s3}(b). 
One object is fixed at one end of the rail, while the other object is released from the same height in each trial. 
The coefficient of restitution is then obtained from the ratio between the outgoing and incoming velocities: $e=v_{\mathrm{out}}/v_{\mathrm{in}}$.

For cloth materials, we use Style3D stretch and bending measurement instruments to characterize stretch and bending stiffness through stretch tests and cantilever bending tests, respectively. 
During stretch measurements, we evaluate deformation along both the warp and weft directions of the fabric, as well as the diagonal direction, to characterize shear stiffness. 
In practical simulation settings, we use the average stretch stiffness along the warp and weft directions as the final parameter configuration.
Finally, to validate the accuracy of our measurements, we compare the real-world and simulated results in a hanging test using a $1,\mathrm{m}^2$ piece of cloth, as shown in \cref{s3}(c).

For deformable materials, we calibrate Young's modulus and Poisson's ratio using compression tests and Digital Image Correlation (DIC), respectively. 
\cref{s3}(d) shows the measured stress-strain curve during material characterization. 
Since our experiments mainly operate in the small-deformation regime, we fit the initial linear region of the curve to estimate Young's modulus. 
\cref{s3}(e) presents the DIC visualization during the measurement process, where Poisson's ratio is obtained from the ratio between transverse expansion and longitudinal elongation.

\subsubsection{Textile parameters mapping }
We denote the measured stretch and shear stiffnesses per unit width by $K_s$ and $K_{\mathrm{sh}}$ (N/m), respectively, the thickness by $t$, and the torque-per-angle stiffness of a discrete bending hinge by $K_\theta$. 
Here, $K_s$ is defined as the tangent stiffness of a uniaxial test with traction-free transverse boundaries. 
Because the evaluated simulators employ different constitutive models and discretizations, the same measurements cannot be transferred as identical numerical coefficients. 
We therefore derive model-specific initializations by matching the corresponding small-strain response. These values are treated as effective isotropic solver parameters rather than exact representations of the generally anisotropic textile response.

\paragraph{IsaacSim}
For an isotropic membrane under plane stress, the free-transverse uniaxial and shear stiffnesses satisfy
\begin{equation}
    K_s = E t,
    \qquad
    K_{\mathrm{sh}} = \frac{E t}{2(1+\nu)} ,
\end{equation}
which gives Young's modulus $E$ and Poisson's ratio $\nu$:
\begin{equation}
    \nu = \frac{K_s}{2K_{\mathrm{sh}}}-1,
    \qquad
    E = \frac{K_s}{t}.
    \label{eq:isaac_inplane_mapping}
\end{equation}
When $\nu$ falls outside the non-auxetic range supported in our configuration, we preserve the measured stretch stiffness and use the
fixed fallback
\begin{equation}
  \nu =
  \begin{cases}
    \nu, & 0 \leq \nu < 0.5,\\
    0.30,      & \nu < 0,\\
    0.45,      & \nu \geq 0.5.
  \end{cases}
\end{equation}
PhysX represents bending through the modulus-like parameter \texttt{surfaceBendStiffness}, whose resulting edge stiffness is proportional to $t^3$ \cite{nvidia2026deformables}. 
For our fixed, regular square grid with diagonal triangulation, we use the mesh-specific initialization
\begin{equation}
    \texttt{surfaceBendStiffness}
    \simeq \frac{K_\theta}{4t^3}.
    \label{eq:isaac_bending_mapping}
\end{equation}
The factor $1/4$ follows from averaging the two edge classes in this triangulation and is not a mesh-independent identity \cite{Grinspun2003}.

\paragraph{Genesis}
Genesis models textiles using discrete distance and dihedral constraints.
Following the nominal XPBD energy convention, we initialize its per-constraint compliance parameters as
\begin{equation}
    \begin{aligned}
        \texttt{stretch\_compliance} &= \frac{1}{K_s},\\
        \texttt{bending\_compliance} &= \frac{1}{K_\theta}.\\
    \end{aligned}
        \label{eq:genesis_mapping}
\end{equation}
These reciprocals should not be interpreted as an exact, mesh-independent conversion from macroscopic textile properties.
Genesis applies the same coefficients to individual mesh edges and hinges, while the effective response also depends on the mesh, substep size, relaxation factors, and number of projection iterations \cite{Mueller2007,Macklin2016,genesis2026source}. 
We therefore keep these settings fixed and report them with the benchmark.
Moreover, the evaluated Genesis textile model provides no independent input channel for $K_{\mathrm{sh}}$.

\paragraph{Newton VBD}
Newton's VBD membrane uses stable Neo-Hookean elasticity, with its input coefficients corresponding to the small-strain Lamé parameters $\mu$ and $\lambda$. 
Internally, the stable Neo-Hookean parameters are $\mu_{\mathrm{NH}}=\mu$ and $\lambda_{\mathrm{NH}}=\lambda+\mu$ \cite{Smith2018,newton2026source}. 
Matching a free-transverse uniaxial test and an engineering-shear test yields
\begin{equation}
K_{\mathrm{sh}}=\mu,
\qquad
K_s=\frac{4\mu(\lambda+\mu)}{\lambda+2\mu}.
\end{equation}
Consequently, the Newton coefficients are initialized by
\begin{equation}
\texttt{tri\_ke}=\widetilde K_{\mathrm{sh}},
\qquad
\texttt{tri\_ka}=
\frac{
2\widetilde K_{\mathrm{sh}}
\left(K_s-2\widetilde K_{\mathrm{sh}}\right)}
{4\widetilde K_{\mathrm{sh}}-K_s},
\label{eq:newton_mapping}
\end{equation}
where
\begin{equation}
  \widetilde K_{\mathrm{sh}} =
    \begin{cases}
      K_{\mathrm{sh}},
      & 2<K_s/K_{\mathrm{sh}}<4,\\
      K_s/2.7,
      & \rm{otherwise}.
    \end{cases}
\end{equation}
The interval above is the positive-$\lambda$ regime of the intrinsic two-dimensional isotropic model, rather than a universal admissibility condition. 
The ratio $2.7$ is a fixed engineering prior used when the anisotropic measurements cannot be represented by this isotropic model; it is not a theoretical constant of Newton or stable Neo-Hookean elasticity. Finally, because Newton's bending kernel uses the effective hinge coefficient $k_e=\texttt{edge\_ke}\,\ell_e$, we set
\begin{equation}
    \texttt{edge\_ke}=\frac{K_\theta}{\bar{\ell}_e},
    \label{eq:newton_bending_mapping}
\end{equation}
where $\bar{\ell}_e$ is the mean rest-edge length of our nearly uniform mesh. 

\section{Physics engines}
\subsection{Metrics}
\label{app:subsec:metrics}

\begin{figure}[!h]
\centering
\captionsetup[subfigure]{justification=centering,singlelinecheck=true}
\def\figSFiveScale{0.8}
\begin{adjustbox}{width=\figSFiveScale\columnwidth}
\begin{minipage}{\columnwidth}
\centering
\begin{subfigure}[t]{0.48\linewidth}
    \centering
    \includegraphics[width=\linewidth]{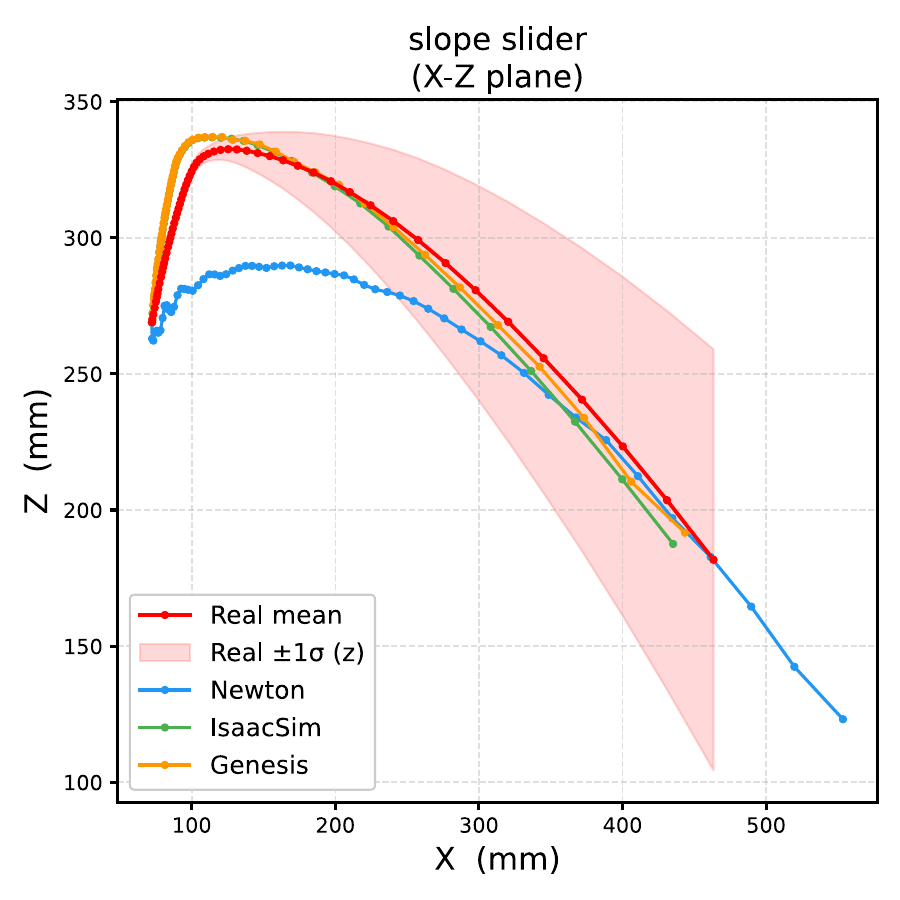}
    \caption{}
    \label{fig:a}
\end{subfigure}
\hfill
\begin{subfigure}[t]{0.48\linewidth}
    \centering
    \includegraphics[width=\linewidth]{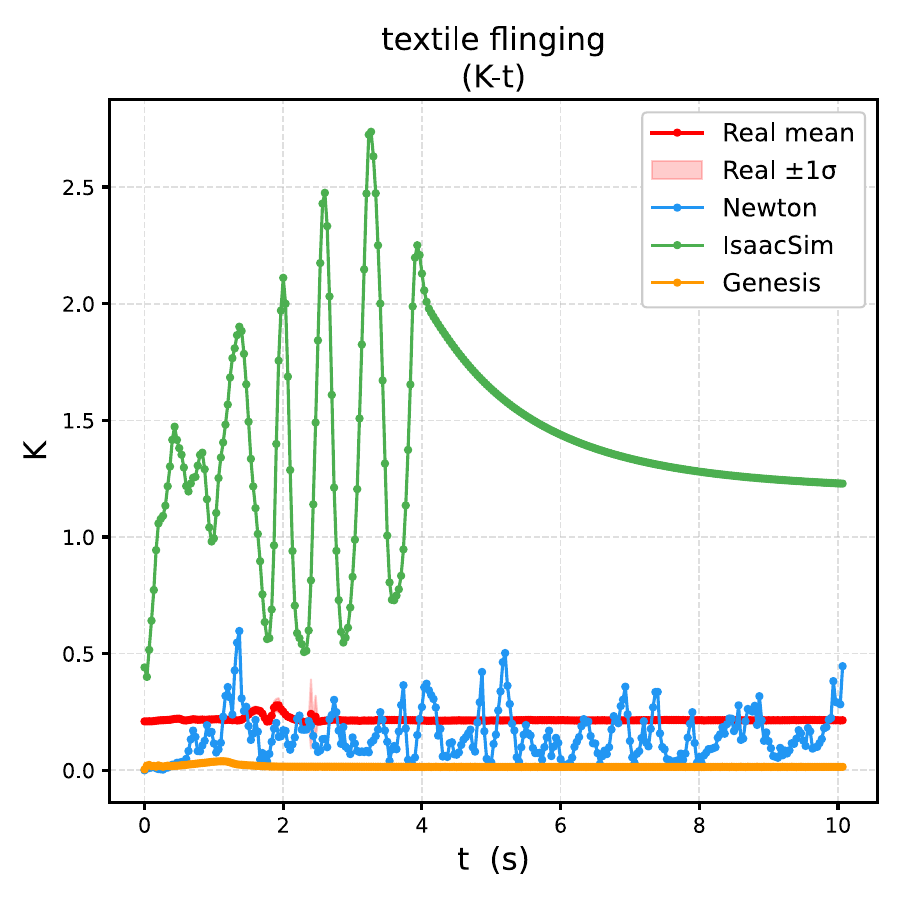}
    \caption{}
    \label{fig:b}
\end{subfigure}
\par\vspace{2mm}
\begin{subfigure}[t]{0.48\linewidth}
    \centering
    \includegraphics[width=\linewidth]{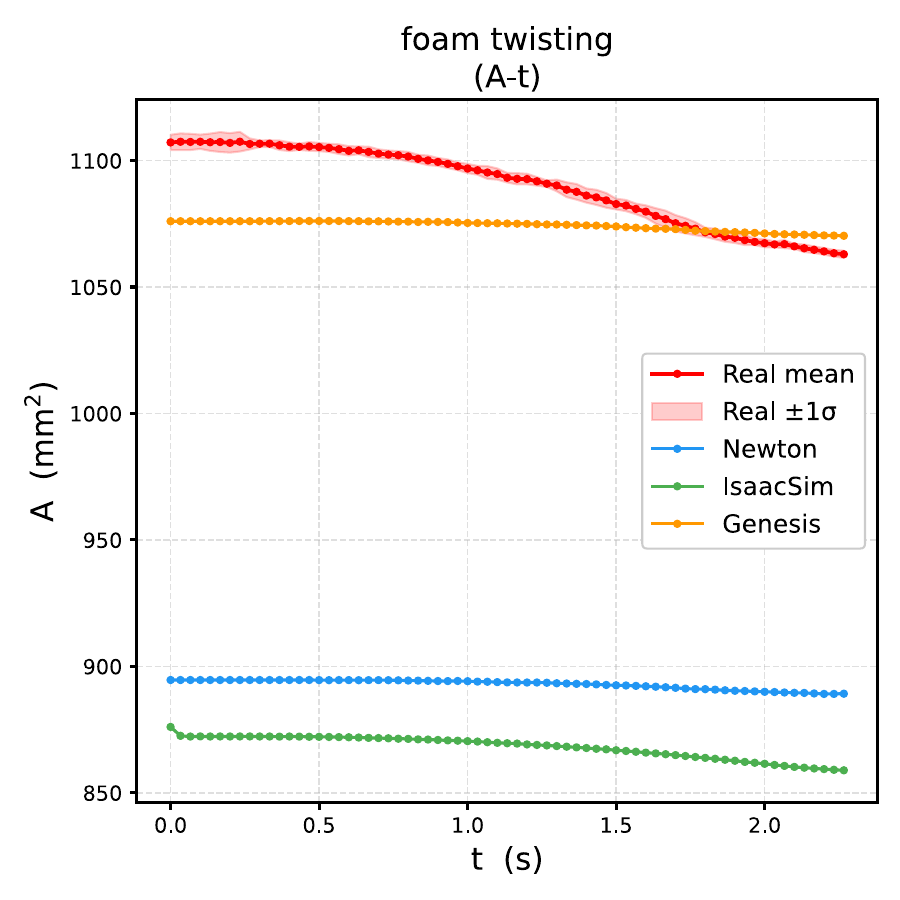}
    \caption{}
    \label{fig:c}
\end{subfigure}
\hfill
\begin{subfigure}[t]{0.48\linewidth}
    \centering
    \includegraphics[width=\linewidth]{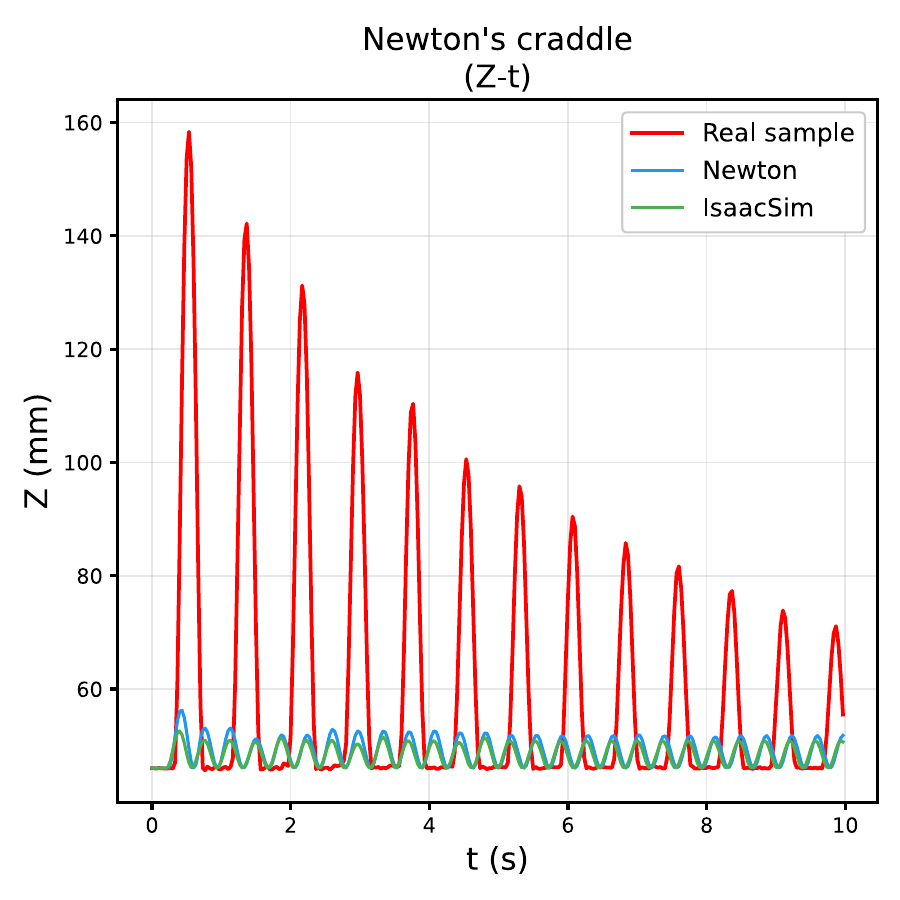}
    \caption{}
    \label{fig:d}
\end{subfigure}
\end{minipage}
\end{adjustbox}
\caption{Representative generalized trajectories used for physics-engine evaluation. Red curves and bands in (a)--(c) show the real-world mean and $\pm1\sigma$ across repeated trials, while the remaining curves show simulator outputs; (d) compares a representative real trial with the available simulations.}
\label{s4}
\end{figure}


\paragraph{Generated trajectory}
As discussed in the main text, when evaluating the performance of the simulation engines, we define a generalized trajectory $g(t)$, which denotes, respectively, the rigid-body position ($P$) in \cref{s4}(a), the Gaussian curvature of the textile marker points ($K$) in \cref{s4}(b), and the area of the triangular faces formed by the soft-body marker points ($A$)in \cref{s4}(c). 
For the Gaussian curvature, the computation adopted in this work is based on the discrete form of the Gauss--Bonnet theorem, namely the \emph{angle deficit method}. 
Specifically, at each time instant, for every marker point $i$ that serves as a vertex of the mesh, the interior angle subtended at vertex $i$ by each adjacent triangle is computed. 
For a given adjacent triangle, let $\mathbf{a}$ and $\mathbf{b}$ denote the edge vectors from vertex $i$ to the other two vertices of that triangle; the corresponding interior angle is
\begin{equation}
    \theta = \arccos\!\left(\frac{\mathbf{a}\cdot\mathbf{b}}{\|\mathbf{a}\|\,\|\mathbf{b}\|}\right).
    \label{eq:interior_angle}
\end{equation}
Summing all interior angles adjacent to vertex $i$, denoted $\sum\theta_i$, the discrete Gaussian curvature at that vertex is defined as the difference between a reference full angle $\Theta_i$ and this angle sum:
\begin{equation}
    k_i = \Theta_i - \sum\theta_i .
    \label{eq:gaussian_curvature}
\end{equation}
Since the fabric samples studied in this work are open, bounded rectangular surface patches rather than closed surfaces, marker points are pre-classified, according to their topological position within the mesh, into three categories: interior points, edge points (lying on a straight boundary), and corner points. 
The reference full angle $\Theta_i$ in \cref{eq:gaussian_curvature} is set accordingly to $2\pi$, $\pi$, and $\pi/2$, respectively, corresponding to the angle subtended when, in the zero-curvature case, an interior point is fully surrounded, an edge point is covered by a half-plane, and a rectangular corner point is covered by a quarter-plane.

This curvature is computed independently, frame by frame, along the entire trajectory. 
If a marker point lacks three-dimensional coordinates in the current frame due to occlusion or other causes, or if an adjacent triangle degenerates owing to coincident vertices, the corresponding interior angle in \cref{eq:gaussian_curvature} is excluded from the summation and does not contribute to the curvature computed for that point at that frame. 
It should be noted that $k_i$ is, in essence, the angular deficit at the vertex (expressed in radians) and is not normalized by the local area associated with the vertex; it therefore does not, strictly speaking, constitute a Gaussian curvature value with the dimension of $\mathrm{length}^{-2}$.
The Gaussian curvature $K$ shown in \cref{s4}(b) is precisely the norm of the vector formed by the per-vertex curvature values $k_i$.

\paragraph{Longest Static Period}
\cref{s4}(d) illustrates a segment of the Newton's cradle trajectory recorded in both the real world and simulation. 
It can be observed that, due to the near-perfect momentum transfer efficiency in the real-world setup, the pendulum balls remain stationary for a certain period after momentum transfer, resulting in an impulse-like trajectory pattern with discrete motion intervals. 
The proposed Longest Static Duration (LSD) metric is specifically designed to identify the stationary periods between consecutive impulses. 
LSD is defined as:
\begin{equation}
\mathrm{LSD}=\max_{S}\{|S|/f_s\}
\quad \mathrm{s.t.}\quad
\begin{cases}
    \max(z_S)-\min(z_S)&\leq\epsilon_z,\\
    \quad |v_S|&\leq v_{\mathrm{th}}\\
\end{cases}
\end{equation}
where $\mathcal{S}$ denotes the set of consecutive frames satisfying the stationary condition, $f_s$ represents the sampling frequency, and $\epsilon_z$ and $v_{\mathrm{th}}$ denote the position variation threshold and velocity threshold, respectively. 
In our evaluation, we set $\epsilon_z=0.5~\mathrm{mm}$ and $v_{\mathrm{th}}=5~\mathrm{mm/s}$.

\paragraph{Momentum transfer efficiency}
Different from LSD, the Momentum Transfer Efficiency (MTE) directly evaluates the efficiency of momentum transmission during collisions from a momentum perspective. 
In our calculation, we only consider the efficiency of the first impact:

\begin{equation}
\mathrm{MTE}=\sqrt{\frac{\Delta H_{\mathrm{out}}}{\Delta H_{\mathrm{in}}}}
\end{equation}
where $\Delta H_{\mathrm{in}}$ and $\Delta H_{\mathrm{out}}$ denote the initial height difference and the maximum swing height difference relative to the equilibrium position of the incoming and outgoing balls, respectively. A larger MTE value indicates that a greater proportion of the incoming ball's momentum is transferred to the outgoing ball.

\paragraph{Period}
The period is a commonly used metric for evaluating simple harmonic motion in pendulum systems. However, since the pendulum in \modelname{} is released from the horizontal position, its motion does not satisfy the analytical oscillation formula derived under the small-angle assumption. 
Therefore, instead of directly applying the theoretical period equation, we first perform Fourier spectral analysis on the trajectory to identify the dominant frequency component. 
The reciprocal of this dominant frequency is then used as the estimated oscillation period of the pendulum.

\paragraph{Energy loss}
The Energy Loss metric is designed to evaluate whether the numerical integrator used by a simulator satisfies the energy conservation principle. It is defined as:
\begin{equation}
\mathrm{EL}=\frac{E_{\mathrm{start}}-E_{\mathrm{end}}}{E_{\mathrm{start}}}
\end{equation}
where $E_{\mathrm{start}}$ and $E_{\mathrm{end}}$ denote the mechanical energy of the system at the initial and final time steps, respectively. 
In practice, we compute this metric at the peak positions of the pendulum oscillation, where the kinetic energy is completely converted into gravitational potential energy, and thus the total mechanical energy can be approximated by the gravitational potential energy.
A value closer to 1 indicates better energy conservation. 
In contrast, a value below 0 indicates an unphysical increase in system energy caused by the accumulation of numerical integration errors.
\cref{tab5} provides detailed evaluation results of \modelname{} across all sub-tasks and object variants.

\begingroup
\small
\setlength{\tabcolsep}{1mm}
\setlength{\LTleft}{\fill}
\setlength{\LTright}{\fill}
\setlength{\LTpre}{0.8\baselineskip}
\setlength{\LTpost}{0.8\baselineskip}
\begin{longtable}{lllllrrrr}
\toprule
{\normalsize\textbf{Task}} &
{\normalsize\textbf{Sub.}} &
{\normalsize\textbf{Mat.}} &
{\normalsize\textbf{Frm.}} &
{\normalsize\textbf{Met.}} &
{\normalsize\textbf{Base.}} &
{\normalsize\textbf{Isaacsim}} &
{\normalsize\textbf{Genesis}} &
{\normalsize\textbf{Newton}} \\
\midrule
\endfirsthead

\multicolumn{9}{c}{{\normalsize \tablename\ \thetable{} -- continued from previous page}} \\
\toprule
{\normalsize\textbf{Task}} &
{\normalsize\textbf{Sub.}} &
{\normalsize\textbf{Mat.}} &
{\normalsize\textbf{Frm.}} &
{\normalsize\textbf{Met.}} &
{\normalsize\textbf{Base.}} &
{\normalsize\textbf{Isaacsim}} &
{\normalsize\textbf{Genesis}} &
{\normalsize\textbf{Newton}} \\
\midrule
\endhead

\midrule
\multicolumn{9}{r}{Continued on next page} \\
\endfoot

\endlastfoot

slope contact
& 
& wood
& 10
& RMSE
& 9.25 (5.17)
& 3.38
& 2.98
& \textbf{2.66} \\

& 
& 
& 
& DTW
& 6.27 (2.48)
& 2.94
& 3.05
& \textbf{2.66} \\

& 
& plastic
& 10
& RMSE
& 12.31 (7.62)
& \textbf{1.26}
& 1.6
& 1.75 \\

& 
& 
& 
& DTW
& 5.30 (1.84)
& \textbf{1.83}
& 2.57
& 3.07 \\

\midrule

nonsmooth contact
& 1
& wood
& 11
& RMSE
& 12.51 (6.77)
& 3.51
& \textbf{3.42}
& 3.61 \\

& 
& 
& 
& DTW
& 6.38 (1.78)
& 3.59
& \textbf{3.29}
& 3.3 \\

& 
& plastic
& 11
& RMSE
& 12.94 (5.91)
& 3.42
& 3.43
& \textbf{3.01} \\

& 
& 
& 
& DTW
& 6.48 (1.76)
& 3.75
& 3.73
& \textbf{2.93} \\

& 2
& wood
& 8
& RMSE
& 14.28 (13.20)
& 2.72
& 2.7
& \textbf{2.69} \\

& 
& 
& 
& DTW
& 7.90 (4.64)
& 3.06
& 2.98
& \textbf{2.96} \\

& 
& plastic
& 9
& RMSE
& 20.35 (9.02)
& 2.16
& 2.8
& \textbf{2.15} \\

& 
& 
& 
& DTW
& 10.54 (2.99)
& 2.27
& 2.76
& \textbf{2.11} \\

& 3
& wood
& 13
& RMSE
& 16.98 (10.77)
& \textbf{2.2}
& 5.84
& 2.23 \\

& 
& 
& 
& DTW
& 8.68 (2.25)
& \textbf{2.44}
& 8.21
& 3.11 \\

& 
& plastic
& 20
& RMSE
& 17.67 (6.19)
& \textbf{1.53}
& 10.04
& 6.05 \\

& 
& 
& 
& DTW
& 8.92 (2.44)
& \textbf{1.9}
& 14.37
& 8.01 \\

\midrule

slope slider
& 1
& wood
& 9
& RMSE
& 0.59 (0.28)
& \textbf{0.019}
& 0.074
& 24.68 \\

& 
& 
& 
& DTW
& 0.59 (0.28)
& \textbf{0.017}
& 0.055
& 14.12 \\

& 
& plastic
& 9
& RMSE
& 1.90 (2.17)
& \textbf{0.013}
& 0.016
& 6.24 \\

& 
& 
& 
& DTW
& 1.90 (2.17)
& \textbf{0.011}
& 0.014
& 4.31 \\

& 
& metal
& 9
& RMSE
& 0.86 (0.69)
& 0.078
& \textbf{0.062}
& 15.08 \\

& 
& 
& 
& DTW
& 0.86 (0.69)
& 0.067
& \textbf{0.052}
& 10.63 \\

& 2
& wood
& 19
& RMSE
& 18.00 (15.89)
& 1.23
& \textbf{0.53}
& 1.71 \\

& 
& 
& 
& DTW
& 7.70 (3.69)
& 1.11
& \textbf{1.05}
& 1.79 \\

& 
& plastic
& 17
& RMSE
& 25.01 (25.45)
& \textbf{0.48}
& 0.82
& 0.51 \\

& 
& 
& 
& DTW
& 9.13 (5.00)
& \textbf{0.87}
& 1
& 1.06 \\

& 
& metal
& 19
& RMSE
& 32.06 (21.93)
& 1
& \textbf{0.83}
& 1.15 \\

& 
& 
& 
& DTW
& 10.87 (4.04)
& \textbf{0.45}
& 0.54
& 0.9 \\

& 3
& wood
& 46
& RMSE
& 24.27 (16.85)
& 2.62
& 2.62
& \textbf{1.35} \\

& 
& 
& 
& DTW
& 6.73 (2.56)
& 2.21
& 2.24
& \textbf{2.16} \\

& 
& plastic
& 46
& RMSE
& 49.96 (20.37)
& 1.58
& 1.46
& \textbf{0.67} \\

& 
& 
& 
& DTW
& 12.97 (4.94)
& 2.2
& 1.83
& \textbf{1.55} \\

& 
& metal
& 60
& RMSE
& 38.01 (28.53)
& 0.61
& \textbf{0.58}
& 1.95 \\

& 
& 
& 
& DTW
& 9.54 (8.52)
& 0.71
& \textbf{0.69}
& 2.93 \\

\midrule

turntable
& 1
& wood
& 95
& RMSE
& 2.01 (1.45)
& \textbf{0.32}
& 0.83
& 34.37 \\

& 
& 
& 
& DTW
& 1.40 (0.64)
& \textbf{0.43}
& 1.14
& 36.33 \\

& 
& plastic
& 95
& RMSE
& 2.00 (2.48)
& 1.22
& \textbf{1.12}
& 36.04 \\

& 
& 
& 
& DTW
& 0.97 (0.61)
& \textbf{2.19}
& 2.22
& 53.08 \\

& 
& metal
& 95
& RMSE
& 3.17 (1.74)
& \textbf{0.42}
& 0.51
& 17.19 \\

& 
& 
& 
& DTW
& 1.53 (0.50)
& \textbf{0.82}
& 0.99
& 28.43 \\

& 2
& wood
& 18
& RMSE
& 6.76 (4.65)
& 4.82
& 3.69
& \textbf{2.88} \\

& 
& 
& 
& DTW
& 4.70 (2.68)
& 4.64
& 3.56
& \textbf{3.26} \\

& 
& plastic
& 17
& RMSE
& 7.86 (5.21)
& 3.61
& \textbf{1.57}
& 1.92 \\

& 
& 
& 
& DTW
& 3.92 (1.58)
& 4.75
& \textbf{1.96}
& 2.2 \\

& 
& metal
& 17
& RMSE
& 5.76 (4.17)
& 5.7
& 3.09
& \textbf{1.97} \\

& 
& 
& 
& DTW
& 3.60 (1.66)
& 6.06
& 3.16
& \textbf{2.84} \\

& 3
& wood
& 95
& RMSE
& 27.15 (27.08)
& \textbf{1.55}
& 1.75
& 9.5 \\

& 
& 
& 
& DTW
& 14.36 (12.90)
& \textbf{0.85}
& 1.17
& 15.09 \\

& 
& plastic
& 95
& RMSE
& 12.50 (6.49)
& \textbf{2.37}
& 2.47
& 21.76 \\

& 
& 
& 
& DTW
& 3.48 (0.84)
& \textbf{1.78}
& 2.04
& 66.49 \\

& 
& metal
& 95
& RMSE
& 13.26 (7.43)
& \textbf{0.17}
& 0.57
& 20.04 \\

& 
& 
& 
& DTW
& 3.33 (0.81)
& \textbf{0.61}
& 2.01
& 66.72 \\

\midrule

bouncing ball
& 
& rubber
& 18
& RMSE
& 5.29 (3.60)
& \textbf{15.63}
& 22.5
& 22.71 \\

& 
& 
& 
& DTW
& 4.23 (2.87)
& \textbf{5.58}
& 14.58
& 14.89 \\

\midrule

newton cradle
& 
& metal
& 1027
& LSD
& 0.38 (0.015)
& 0
& \textemdash{}
& 0 \\

& 
& 
& 
& MTE
& 93.02 (2.47)
& 0.2
& \textemdash{}
& \textbf{0.26} \\

\midrule

pendulum
& 
& metal
& 300
& PD
& 1.14 (0.066)
& \textbf{1.1}
& 2.47
& 1.09 \\

& 
& 
& 3000
& EL
& 0.00
& 0.034
& \textbf{-0.041}
& 0.022 \\

\midrule

textile stretching
& 
& rayon
& 432
& RMSE
& 0.024 (0.030)
& 7.43
& 12.15
& \textbf{7.34} \\

& 
& 
& 
& DTW
& 0.0084 (0.0018)
& 21.06
& 34.49
& \textbf{20.8} \\

& 
& satin
& 380
& RMSE
& 0.21 (0.42)
& \textbf{0.73}
& 1.51
& \textbf{0.73} \\

& 
& 
& 
& DTW
& 0.21 (0.42)
& \textbf{0.76}
& 1.56
& 0.75 \\

& 
& uniform cloth
& 432
& RMSE
& 0.018 (0.012)
& \textbf{6.12}
& 13.79
& \textemdash{} \\

& 
& 
& 
& DTW
& 0.011 (0.0018)
& \textbf{10.16}
& 23.04
& \textemdash{} \\

& 
& nylon taslan
& 448
& RMSE
& 0.027 (0.0077)
& 5.9
& 12.74
& \textbf{5.7} \\

& 
& 
& 
& DTW
& 0.024 (0.0055)
& 6.63
& 14.46
& \textbf{6.44} \\

& 
& Oxford fabric
& 423
& RMSE
& 0.0061 (0.0080)
& 18.05
& 51.16
& \textbf{15.93} \\

& 
& 
& 
& DTW
& 0.0051 (0.0068)
& 21.1
& 61.17
& \textbf{18.9} \\

& 
& synthetic leather
& 441
& RMSE
& 0.017 (0.0072)
& \textbf{6.12}
& 18.21
& \textemdash{} \\

& 
& 
& 
& DTW
& 0.0085 (0.0038)
& \textbf{11.97}
& 37.2
& \textemdash{} \\

\midrule

textile bending
& 
& rayon
& 379
& RMSE
& 1.07 (0.21)
& \textbf{7.69}
& 20.04
& 37.13 \\

& 
& 
& 
& DTW
& 0.63 (0.13)
& \textbf{11.67}
& 31.9
& 58.97 \\

& 
& satin
& 335
& RMSE
& 1.92 (0.69)
& \textbf{7.94}
& 11.73
& 19.9 \\

& 
& 
& 
& DTW
& 1.20 (0.17)
& \textbf{11.78}
& 17.35
& 18.82 \\

& 
& uniform cloth
& 302
& RMSE
& 1.28 (0.34)
& \textbf{7.63}
& 14.15
& 36.77 \\

& 
& 
& 
& DTW
& 0.48 (0.058)
& \textbf{17.88}
& 35.05
& 90 \\

& 
& nylon taslan
& 187
& RMSE
& 1.94 (0.95)
& \textbf{7.98}
& 9.97
& 20.02 \\

& 
& 
& 
& DTW
& 1.37 (0.94)
& \textbf{9.57}
& 12.57
& 25.19 \\

& 
& Oxford fabric
& 145
& RMSE
& 2.47 (1.26)
& \textbf{4.78}
& 7.84
& 12.53 \\

& 
& 
& 
& DTW
& 1.96 (0.84)
& \textbf{5.29}
& 8.88
& 13.6 \\

& 
& synthetic leather
& 302
& RMSE
& 1.79 (0.45)
& \textbf{12.49}
& 14.63
& 21.79 \\

& 
& 
& 
& DTW
& 1.23 (0.43)
& \textbf{16.98}
& 19.98
& 21.19 \\

\midrule

textile flinging
& 
& rayon
& 303
& RMSE
& 0.024 (0.0087)
& 61.62
& \textbf{9.31}
& 11.09 \\

& 
& 
& 
& DTW
& 0.016 (0.0015)
& 91.62
& \textbf{14.36}
& 16.66 \\

& 
& satin
& 562
& RMSE
& 0.016 (0.0056)
& 128.26
& \textbf{8.54}
& 9.25 \\

& 
& 
& 
& DTW
& 0.012 (0.0014)
& 167.31
& \textbf{11.27}
& 12.12 \\

& 
& uniform cloth
& 444
& RMSE
& 0.016 (0.0067)
& 78.06
& \textbf{7.11}
& 18.35 \\

& 
& 
& 
& DTW
& 0.012 (0.0019)
& 105.17
& \textbf{9.77}
& 22.22 \\

& 
& nylon taslan
& 204
& RMSE
& 0.021 (0.020)
& 104.87
& \textbf{3.66}
& 11.13 \\

& 
& 
& 
& DTW
& 0.0095 (0.0022)
& 222.9
& \textbf{7.94}
& 20.56 \\

& 
& Oxford fabric
& 187
& RMSE
& 0.013 (0.0097)
& 115.3
& \textbf{5.3}
& 15.46 \\

& 
& 
& 
& DTW
& 0.0095 (0.0014)
& 148.27
& \textbf{7.22}
& 17.85 \\

& 
& synthetic leather
& 219
& RMSE
& 0.0072 (0.00051)
& 77.49
& \textbf{6.91}
& 27.65 \\

& 
& 
& 
& DTW
& 0.0063 (0.00023)
& 84.39
& \textbf{7.89}
& 25.72 \\

\midrule

foam stretching
& 
& Soft
& 30
& RMSE
& 8.72 (6.94)
& 16.15
& \textbf{10.05}
& 15.46 \\

& 
& 
& 
& DTW
& 8.19 (6.93)
& 17.18
& \textbf{10.65}
& 16.45 \\

& 
& hard
& 30
& RMSE
& 21.28 (12.52)
& 11.29
& \textbf{1.74}
& 10.75 \\

& 
& 
& 
& DTW
& 21.07 (12.35)
& 11.41
& \textbf{1.75}
& 10.86 \\

\midrule

foam shearing
& 
& Soft
& 37
& RMSE
& 5.60 (0.70)
& 26.13
& \textbf{15.26}
& 26.57 \\

& 
& 
& 
& DTW
& 5.07 (0.59)
& 28.85
& \textbf{16.74}
& 29.34 \\

& 
& hard
& 38
& RMSE
& 6.25 (3.53)
& 38.57
& \textbf{6.12}
& 36.67 \\

& 
& 
& 
& DTW
& 5.08 (2.01)
& 47.46
& \textbf{7.53}
& 45.12 \\

\midrule

foam twisting
& 
& Soft
& 65
& RMSE
& 8.07 (0.53)
& 17.14
& \textbf{10.86}
& 16.78 \\

& 
& 
& 
& DTW
& 7.48 (0.39)
& 18.47
& \textbf{11.67}
& 18.07 \\

& 
& hard
& 69
& RMSE
& 7.08 (2.63)
& 32.41
& \textbf{4.7}
& 30.12 \\

& 
& 
& 
& DTW
& 5.64 (1.18)
& 40.66
& \textbf{5.86}
& 37.76 \\

\midrule

foam bending
& 
& Soft
& 35
& RMSE
& 7.00 (4.39)
& 10.71
& 14.43
& \textbf{10.37} \\

& 
& 
& 
& DTW
& 3.52 (1.05)
& 21.28
& 27.51
& \textbf{20.53} \\

& 
& hard
& 33
& RMSE
& 2.85 (0.52)
& 58.39
& \textbf{9.96}
& 55.78 \\

& 
& 
& 
& DTW
& 2.28 (0.21)
& 72.88
& \textbf{12.41}
& 69.62 \\
\bottomrule
\caption{Simulation comparison results across rigid-body, textile, and foam tasks. RMSE and DTW report trajectory discrepancies; LSD, MTE, PD, and EL denote the task-specific evaluation metrics. Lower is better unless otherwise specified. Bold marks the best valid simulator result, and dashes denote unavailable results. Abbreviations: Sub. = subtask, Mat. = material, Frm. = frames, Met. = metric, and Base. = baseline.}
\label{tab5} \\
\end{longtable}
\endgroup

\section{World model}
\subsection{Model configurations}
\subsubsection{Standard input}
\label{app:subsubsec:standard input}
\paragraph{First frame}
For the five image-to-video models, every source first frame is converted to a common 832 × 480-pixel canvas before inference. 
Let the original image width and height be W and H. 
The isotropic scale factor is $s = \min(\frac{832}{W}, \frac{480}{H})$.
The image is resized to integer dimensions corresponding to sW and sH, preserving the original aspect ratio and full field of view. 
The resized image is then centered on an 832 × 480 black canvas with symmetric padding. 
No content is cropped and no geometric distortion is introduced. 
Genie 3 is treated as an interface-specific exception because Project Genie requires an input image of at least 1280 × 704 pixels. 
Its first frame is therefore prepared at 1300 × 750, which has the same aspect ratio as 832 × 480 and preserves the standardized scene composition while satisfying the larger input requirement. 
The exported Genie 3 video has a resolution of 1280 × 704. Details can be found in the table below~\cref{tab6}.

Negative prompts are passed through model-specific interfaces. 
For the two Cosmos3 models, the negative prompt is supplied through the Diffusers negative\_prompt argument. 
For Wan-2.2, it is written to wan\_shared\_cfg.sample\_neg\_prompt in shared\_config; this variable is set to an empty string in the baseline condition. 
Wan-2.7 accepts a negative prompt through its API. 
The Seedance 2.0 and Genie 3 interfaces used in this study do not expose a separate negative-prompt field.

\begin{table*}[!h]
    \centering
    \label{tab:model-configurations}
    \renewcommand{\arraystretch}{1.15}
    \setlength{\tabcolsep}{5pt}

    \begin{tabularx}{\textwidth}{
        @{}
        l
        >{\raggedright\arraybackslash}X
        >{\centering\arraybackslash}X
        >{\centering\arraybackslash}X
        @{}
    }
        \toprule
        \textbf{Model}
        &
        \textbf{Access method}
        &
        \textbf{Generated MP4}
        &
        \textbf{Deployment environment}
        \\
        \midrule

        Cosmos3-Nano
        &
        Diffusers \texttt{Cosmos3OmniPipeline}
        &
        \makecell{832 $\times$ 480\\189 frames; 24 fps}
        &
        \makecell{1 $\times$ RTX 4090\\DSW or DLC}
        \\

        Cosmos3-Super-I2V
        &
        Diffusers \texttt{Cosmos3OmniPipeline}
        &
        \makecell{832 $\times$ 480\\189 frames; 24 fps}
        &
        \makecell{8 $\times$ RTX 4090\\DLC}
        \\

        Wan-2.2-I2V-A14B
        &
        Official inference script \texttt{generate.py}
        &
        \makecell{832 $\times$ 464\\81 frames; 16 fps}
        &
        \makecell{8 $\times$ RTX 4090\\DLC}
        \\

        Wan-2.7
        &
        DashScope API
        &
        \makecell{1264 $\times$ 728\\240 frames; 30 fps}
        &
        Remote server
        \\

        Seedance 2.0
        &
        Volcano Engine Ark API
        &
        \makecell{864 $\times$ 496\\193 frames; 24 fps}
        &
        Remote server
        \\

        Genie 3
        &
        Project Genie interface
        &
        \makecell{1280 $\times$ 704\\200 frames; 20 fps}
        &
        Remote server
        \\

        \bottomrule
    \end{tabularx}
    \caption{
        Access methods, generated-video settings, and deployment environments
        of the evaluated world models.
    }
    \label{tab6}
\end{table*}

\paragraph{Text prompts}
The benchmark covers seven physical task families: slope contact, non-smooth contact, slope slider, turntable, Newton's cradle, pendulum, and bouncing ball. 
Material variants are introduced for tasks whose friction, restitution, inertia, or contact behavior depends on the object material. 
The same task-specific prompts are used consistently across all evaluated world models:

\textbf{Slope Contact}
\begin{promptbox}
\textbf{Prompt.}
Static fontal camera view: An 84g solid wood trirectangular tetrahedron is placed on a 997g fixed wood base on a black perforated optical table, initially forming a single upright rectangular block with a diagonal contact seam. 
Governed by gravity, a 0.39 wood-on-wood friction coefficient, and a 0.37 restitution coefficient, the tetrahedron slides down from rest from the slope surface, and bounces on the table until settling. 
Maintain structural integrity and object consistency throughout: no changing shapes, no blending, and no flashing background.
\end{promptbox}

\textbf{Non-smooth Contact (plastic)}
\begin{promptbox}
\textbf{Prompt.}
Static frontal camera view: A stationary robot arm holds a 98g black plastic pyramid (apex oriented downward with a slight tilt) above a fixed 115g wooden square base with a square recess on a black optical table. 
The gripper opens, releasing the wood pyramid into a gravity-driven free fall. 
Governed by a 0.28 plastic-on-wood friction coefficient and an about 0.36 restitution coefficient, the pyramid descends, collides with the base, and bounces until settling into a stable rest, while the robot arm remains completely fixed. 
Maintain structural integrity and object consistency throughout: no changing shapes, no blending, and no flashing background.
\end{promptbox}

\textbf{Non-smooth Contact (wood)}
\begin{promptbox}
\textbf{Prompt.}
Static frontal camera view: A stationary robot arm holds a 43g wooden pyramid (apex oriented downward with a slight tilt) above a fixed 115g wooden square base with a square recess on a black optical table. 
The gripper opens, releasing the wood pyramid into a gravity-driven free fall. 
Governed by a 0.27 wood-on-wood friction coefficient and an about 0.33 restitution coefficient, the pyramid descends, collides with the base, and bounces until settling into a stable rest, while the robot arm remains completely fixed. 
Maintain structural integrity and object consistency throughout: no changing shapes, no blending, and no flashing background.
\end{promptbox}

\textbf{Slope Slider (plastic)}
\begin{promptbox}
\textbf{Prompt.}
Static frontal camera view: A 147g black plastic cube sits near the upper end of a 477g fixed wooden plank. The steep incline of the blank is 30 degree. 
A linear ruler and digital protractor are placed beside the assembly, all resting on a black perforated optical table. 
Coefficient of friction between plastic and wood is 0.28; coefficient of restitution for the cube is 0.38. 
Upon release, the cube slides down the inclined surface, and finally lands on the table surface. 
The wooden plank remains fully stationary for the entire duration of the motion. 
Maintain structural integrity and object consistency throughout: no changing shapes, no blending, and no flashing background.
\end{promptbox}

\textbf{Slope Slider (wood)}
\begin{promptbox}
\textbf{Prompt.}
Static frontal camera view: A 90g wood cube sits near the upper end of a 477g fixed wooden plank. 
The steep incline of the blank is 30 degree. 
A linear ruler and digital protractor are placed beside the assembly, all resting on a black perforated optical table. 
Coefficient of friction between wood and wood is 0.27; coefficient of restitution for the cube is 0.22. 
Upon release, the cube slides down the inclined surface, and finally lands on the table surface. 
The wooden plank remains fully stationary for the entire duration of the motion. 
Maintain structural integrity and object consistency throughout: no changing shapes, no blending, and no flashing background.
\end{promptbox}

\textbf{Slope Slider (metal)}
\begin{promptbox}
\textbf{Prompt.}
Static frontal camera view: A 340g silver metal cube sits near the upper end of a 477g fixed wooden plank. 
The steep incline of the blank is 30 degree. 
A linear ruler and digital protractor are placed beside the assembly, all resting on a black perforated optical table. 
Coefficient of friction between plastic and wood is 0.26; coefficient of restitution for the cube is 0.33. 
Upon release, the cube slides down the inclined surface, and finally lands on the table surface. 
The wooden plank remains fully stationary for the entire duration of the motion. 
Maintain structural integrity and object consistency throughout: no changing shapes, no blending, and no flashing background.
\end{promptbox}

\textbf{Turntable (plastic)}
\begin{promptbox}
\textbf{Prompt.}
Static top camera view: A 147g black plastic cube is positioned off-center atop a 722g wooden circular turntable driven by an electric motor undergoing a counterclockwise rotation with a constant angular velocity; the full assembly rests on an optical table. 
Coefficient of friction between the cube and turntable is 0.32; coefficient of restitution is approximately 0.32. 
During rotation, the wood cube slides on the turntable surface. 
Maintain structural integrity and object consistency throughout: no changing shapes, no blending, and no flashing background.
\end{promptbox}

\textbf{Turntable (wood)}
\begin{promptbox}
\textbf{Prompt.}
Static top camera view: A 90g wood cube is positioned off-center atop a 722g wooden circular turntable driven by an electric motor undergoing a counterclockwise rotation with a constant angular velocity; the full assembly rests on an optical table. 
Coefficient of friction between the cube and turntable is 0.38; coefficient of restitution is 0.29. 
During rotation, the wood cube slides on the turntable surface. 
Maintain structural integrity and object consistency throughout: no changing shapes, no blending, and no flashing background.
\end{promptbox}

\textbf{Turntable (metal)}
\begin{promptbox}
\textbf{Prompt.}
Static top camera view: A 340g silver metal cube is positioned off-center atop a 722g wooden circular turntable driven by an electric motor undergoing a counterclockwise rotation with a constant angular velocity; the full assembly rests on an optical table. 
Coefficient of friction between the cube and turntable is 0.33; coefficient of restitution is approximately 0.23. 
During rotation, the wood cube slides on the turntable surface. 
Maintain structural integrity and object consistency throughout: no changing shapes, no blending, and no flashing background.
\end{promptbox}

\textbf{Newton’s Cradle}
\begin{promptbox}
\textbf{Prompt.}
Static frontal camera view: A Newton's cradle assembly is consist of five identical 33g metal balls, and rests atop a black perforated optical table. 
Four balls hang in contact forming a stationary aligned row; the leftmost ball is pulled back to a nearly horizontal position on its individual suspension string and held propped against the tip of an independent lever mounted to a sliding rail. 
Upon release, the ball swings back to collide with the stationary row. 
Momentum transfers sequentially through the chain of balls, launching the terminal ball outward into a swing. 
The collisions are moderately inelastic, which visibly suppresses the ideal Newton's cradle behavior across successive oscillation cycles. 
The coefficient of restitution between balls is 0.20. 
Maintain structural integrity and object consistency throughout: no changing shapes, no blending, and no flashing background.
\end{promptbox}

\textbf{Pendulum}
\begin{promptbox}
\textbf{Prompt.}
Static frontal camera view: A 318g black metal ball is fixed to the tip of a lightweight rigid rod, which rotates freely about a stationary horizontal axle mounted between two vertical posts atop a black perforated optical table. 
Initially, the rod angle is 10 degree, with the ball extending out to one side. 
Upon release, the ball follows pendulum motion: it accelerates downward past the black vertical posts, ascends on the opposite side, and continues oscillating back and forth over numerous cycles, its amplitude slowly and gradually diminishing due to friction. 
Maintain structural integrity and object consistency throughout: no changing shapes, no blending, and no flashing background.
\end{promptbox}

\textbf{Bouncing Ball}
\begin{promptbox}
\textbf{Prompt.}
Static frontal camera view: A 121g black ball is suspended high above a wooden plank, which is fixed on a black perforated optical table. 
Coefficient of friction is 0.81; coefficient of restitution is 0.58. 
When released, the ball undergoes free fall, followed by a sequence of successive bounces with diminishing peak heights, before finally coming to rest on the plank. 
Maintain structural integrity and object consistency throughout: no changing shapes, no blending, and no flashing background.
\end{promptbox}

\textbf{Negative Prompts}
Cosmos3-Super provides a structured negative\_prompt.json containing separate descriptions for undesirable subjects, spatial relations, backgrounds, lighting, aesthetics, cinematography, temporal consistency, and physical realism. 
The original physical\_realism field is reproduced below.
"No adherence to physical laws. 
Objects defy gravity, pass through solid surfaces, and change mass and momentum without cause. 
Fluid dynamics, cloth simulation, and rigid body physics are all fundamentally broken. 
Furthermore, conservation of energy is violated as objects gain or lose kinetic energy spontaneously. 
Elastic collisions produce inelastic results and vice versa. 
Surface friction is inconsistent -- objects slide on rough surfaces while sticking to smooth ones. 
Air resistance appears to affect only some objects while others move through the atmosphere unimpeded."
To satisfy the character limit of the Wan interfaces while retaining the principal physical failure modes, this field is compressed to the following 493-character prompt.

\begin{negpromptbox}
\textbf{Physical Realism Negative Prompt.}
No adherence to physical laws. 
Objects defy gravity, pass through solid surfaces, and change mass and momentum without cause. 
Broken fluid dynamics, cloth simulation, rigid-body physics and conservation of energy; objects gain or lose kinetic energy spontaneously. 
Elastic collisions produce inelastic results and vice versa. 
Surface friction is inconsistent: objects slide on rough surfaces or stick to smooth ones. 
Air resistance affects some objects while others move through air unimpeded.
\end{negpromptbox}

The same compact negative prompt is applied to Cosmos3-Nano, Cosmos3-Super-I2V, Wan-2.2, and Wan-2.7. 
It is passed through shared\_config for Wan-2.2, through the Diffusers pipeline for the two Cosmos3 models, and through the API negative-prompt field for Wan-2.7. Seedance 2.0 and Genie 3 are excluded from this condition because the interfaces used in this study do not provide a separate negative-prompt input.
To ensure compatibility with the Wan interfaces and maintain a shared negative prompt, the original entry was condensed to 493 characters while preserving its principal physical failure modes.

\subsection{Trajectory recording}
\subsubsection{SAM3}
\label{app:subsubsec:SAM3}
The videos generated in the first stage are subsequently passed to a second-stage segmentation and tracking pipeline based on Segment Anything with Concepts (SAM3). 
SAM3 is a promptable image and video segmentation model that accepts both textual descriptions and visual prompts, such as point prompts, and propagates the selected object mask throughout a video sequence.

\begin{figure*}[!htp]
\centering
\includegraphics[width=0.7\textwidth]{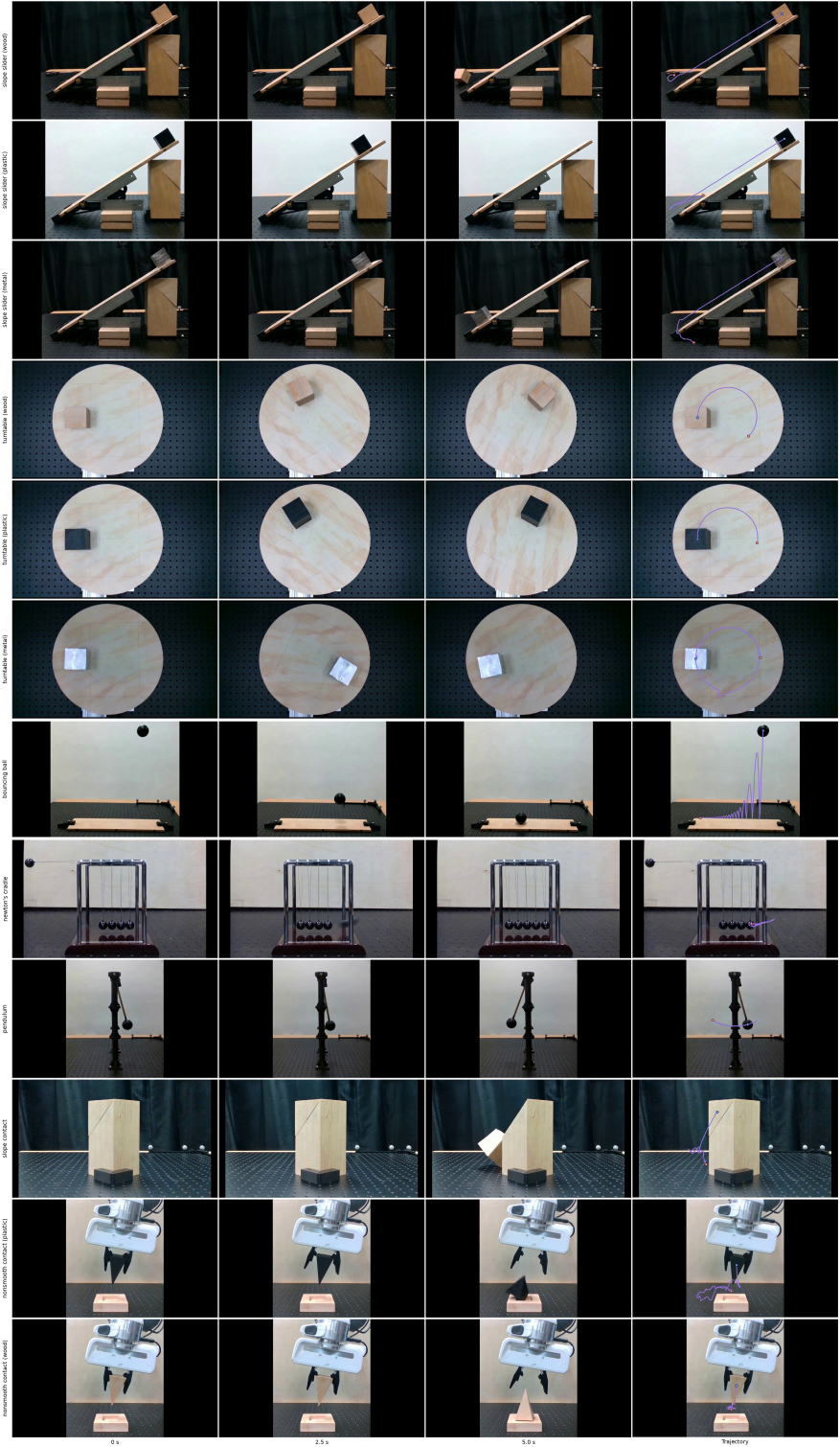}
\caption{Example of qualitative trajectory visualization for physical realism evaluation. The last three rows of the figure correspond to the inference results of "slope contact" and "non-smooth contact". Since these scenarios exhibit substantial 3D geometric deformations and discontinuous object states, they are excluded from the subsequent quantitative evaluation.
}
\label{s5}
\end{figure*}

For each physical task, a reference text prompt and a set of point prompts are initialized on the standardized first frame. 
Before processing the generated videos, the reference point coordinates, text description, and preprocessing metadata are recorded. 
Although the image-to-video models receive the same standardized $832 \times 480$ first frame, their generated videos do not necessarily preserve the same spatial resolution or reproduce the input geometry exactly at the pixel level. 
Differences in native output resolution, internal resizing, retained padding, first-frame reconstruction, and API-side preprocessing may shift the target relative to the canonical annotation. 
Therefore, the point-prompt coordinates are recalibrated separately for the first frame of each generated video.

For outputs that retain the $832 \times 480$ format, the recalibrated points typically remain close to the canonical coordinates. 
For videos generated at other resolutions, the coordinates are first transformed according to the corresponding horizontal and vertical scale factors and are then adjusted using output-specific offsets to account for padding or composition changes. 
After the target location has been verified, the final text prompt, point prompts, frame count, frames per second (FPS), video resolution, and coordinate-transformation parameters are stored in JSON files.

SAM3 is then applied to every generated video. 
For each frame $t$, the model produces a binary segmentation mask for the target object. 
If a predicted mask contains no foreground pixels, the corresponding frame is retained as a missing tracking observation rather than being assigned an artificial object position.

For a valid mask containing $N$ foreground pixels with coordinates
$\{(x_i,y_i)\}_{i=1}^{N}$, the geometric centroid is defined as
\begin{equation}
c_x(t)=\frac{1}{N}\sum_{i=1}^{N}x_i,
\qquad
c_y(t)=\frac{1}{N}\sum_{i=1}^{N}y_i.
\end{equation}
Consistent with the implemented extraction procedure, the mean coordinates are converted to integer pixel positions, and each valid trajectory observation is stored as
\begin{equation}
\left(t,c_x(t),c_y(t)\right),
\end{equation}
where $t$ denotes the frame index. 
After trajectory extraction, the resulting trajectories are then passed to the downstream physical-consistency analysis.

\subsubsection{Results Display}
To illustrate the qualitative evaluation protocol, we visualize one representative example generated by Seedance 2.0. 
For each scenario, frames at 0 s, 2.5 s, and 5.0 s are presented together with the extracted object trajectory. 
The trajectory is obtained by SAM3-based segmentation and centroid tracking, where the start and end positions indicate the tracked motion range. 
Results can be found in~\cref{s5}.

\subsection{Physical-Consistency Analysis}
\label{app:subsec:physical analysis}
This section details how the physical-consistency metrics in the main-paper world-model table are computed.  The tasks are presented in the same order as in that table: slope slider, turntable, bouncing ball, Newton's cradle, and pendulum.  
The analysis uses object-centroid trajectories recovered by segmentation and tracking.  
Unless explicitly stated otherwise, the retained coordinates are neither smoothed nor interpolated, so tracking jumps, stalls, reversals, and model-generated geometric drift remain visible in the reported metrics.

\subsubsection{Common Coordinate and Time Processing}
\label{app:common-processing}
For a generated video with width \(W\), height \(H\), and frame rate \(\mathrm{fps}\), the tracked image coordinate at frame \(f_i\) is \(\mathbf p_i=(x_i,y_i)\).  Because the models generate different native resolutions, coordinates are first mapped to the task's reference image of size \(W_{\mathrm{ref}}\times H_{\mathrm{ref}}\):
\begin{equation}
    \widetilde{x}_i=x_i\frac{W_{\mathrm{ref}}}{W},
    \qquad
    \widetilde{y}_i=y_i\frac{H_{\mathrm{ref}}}{H}.
    \label{eq:coordinate-rescaling}
\end{equation}
Horizontal and vertical scaling are performed independently.  Each task then uses an object of known size to obtain a metric scale \(c\), in \(\mathrm{m/pixel}\), and converts the reference-image coordinates to metric coordinates.  
Time is computed from the original JSON frame keys rather than from the position of a sample in an array:
\begin{equation}
    t_i=\frac{f_i-f_0}{\mathrm{fps}},
    \label{eq:frame-time}
\end{equation}
where \(f_0\) is the first frame retained for the task-specific fit.  
Thus, missing frame keys preserve their true temporal gap.  
Leading stationary frames may be excluded when they precede the physical event, but motion artifacts inside the evaluated interval are retained.

\subsubsection{Slope Slider}
\label{app:c3}

\paragraph{Trajectory Calibration and Projection}
This task tests whether a block released from rest follows uniformly accelerated motion along a fixed incline.  A visible face of the \(L_{\mathrm{block}}=0.05\,\mathrm m\) block is annotated in the reference frame.  
If \(\overline L_{\mathrm{px}}\) is the mean pixel length of its four annotated edges, the metric scale is
\begin{equation}
    c=\frac{L_{\mathrm{block}}}{\overline L_{\mathrm{px}}}.
    \label{eq:c3-scale}
\end{equation}
Two annotated incline points define a fixed unit direction \(\mathbf e\).
After applying \cref{eq:coordinate-rescaling} and the scale \(c\), the signed displacement along the incline is
\begin{equation}
    s_i=\left(\mathbf p_i^{\mathrm m}-\mathbf p_0^{\mathrm m}\right)
    \mathbin{\cdot}\mathbf e.
    \label{eq:c3-projection}
\end{equation}
The incline direction is not re-estimated from each generated trajectory; otherwise a nonphysical trajectory could redefine the direction against which it is evaluated.

The analysis removes a sustained leading or trailing stationary run using a six-frame displacement window.  
After forward motion has been established, the first sustained plateau or reversal terminates the fit because it no longer belongs to the first continuous slide.  
These windows select the frame range only; the selected coordinates themselves remain unfiltered.

\paragraph{Acceleration}
For release from rest with constant acceleration,
\begin{equation}
    s(t)=s_0+\frac{1}{2}at^2.
    \label{eq:c3-kinematics}
\end{equation}
Let \(u_i=t_i^2\).  
Ordinary least squares is used to fit
\begin{equation}
    \widehat{s}_i=b+ku_i,
    \qquad
    a=2k.
\label{eq:c3-acceleration}
\end{equation}
The acceleration \(a\), reported in \(\mathrm{m/s^2}\), measures the physical time scale implied by the generated trajectory.  
It is a parameter estimate, not a goodness-of-fit score, and is therefore compared with the measured real-world baseline for each material.

\paragraph{Quadratic Form Improvement}
The main table's Quadratic Form Improvement (QFI) tests whether the expected linear relation between \(s\) and \(u=t^2\) is sufficient.  
The linear model in \cref{eq:c3-acceleration} is compared with
\begin{equation}
    \widehat{s}^{\,Q}_i=b_Q+k_Qu_i+c_Qu_i^2
    =b_Q+k_Qt_i^2+c_Qt_i^4.
    \label{eq:c3-quadratic}
\end{equation}
With
\begin{equation}
    \mathrm{SSE}_{L}=\sum_{i=1}^{N}(s_i-\widehat{s}_i)^2,
    \qquad
    \mathrm{SSE}_{Q}=\sum_{i=1}^{N}(s_i-\widehat{s}^{\,Q}_i)^2,
    \label{eq:c3-sse}
\end{equation}
the value reported as QFI is the nested-model \(F\) statistic
\begin{equation}
    \mathrm{QFI}
    =
    \frac{(\mathrm{SSE}_{L}-\mathrm{SSE}_{Q})/1}
    {\mathrm{SSE}_{Q}/(N-3)}.
    \label{eq:qfi}
\end{equation}
Low QFI means that adding the \(t^4\) term provides little additional explanatory power, as expected for ideal constant acceleration.  
High QFI indicates systematic curvature in the \(s\)-versus-\(t^2\) relation.  
Because an \(F\) statistic also depends on sample count, it is interpreted together with the plotted residual pattern rather than as an absolute trajectory error.

\begin{figure*}[!htp]
\centering
\begin{minipage}[t]{0.32\textwidth}
  \centering
  \includegraphics[width=\linewidth]{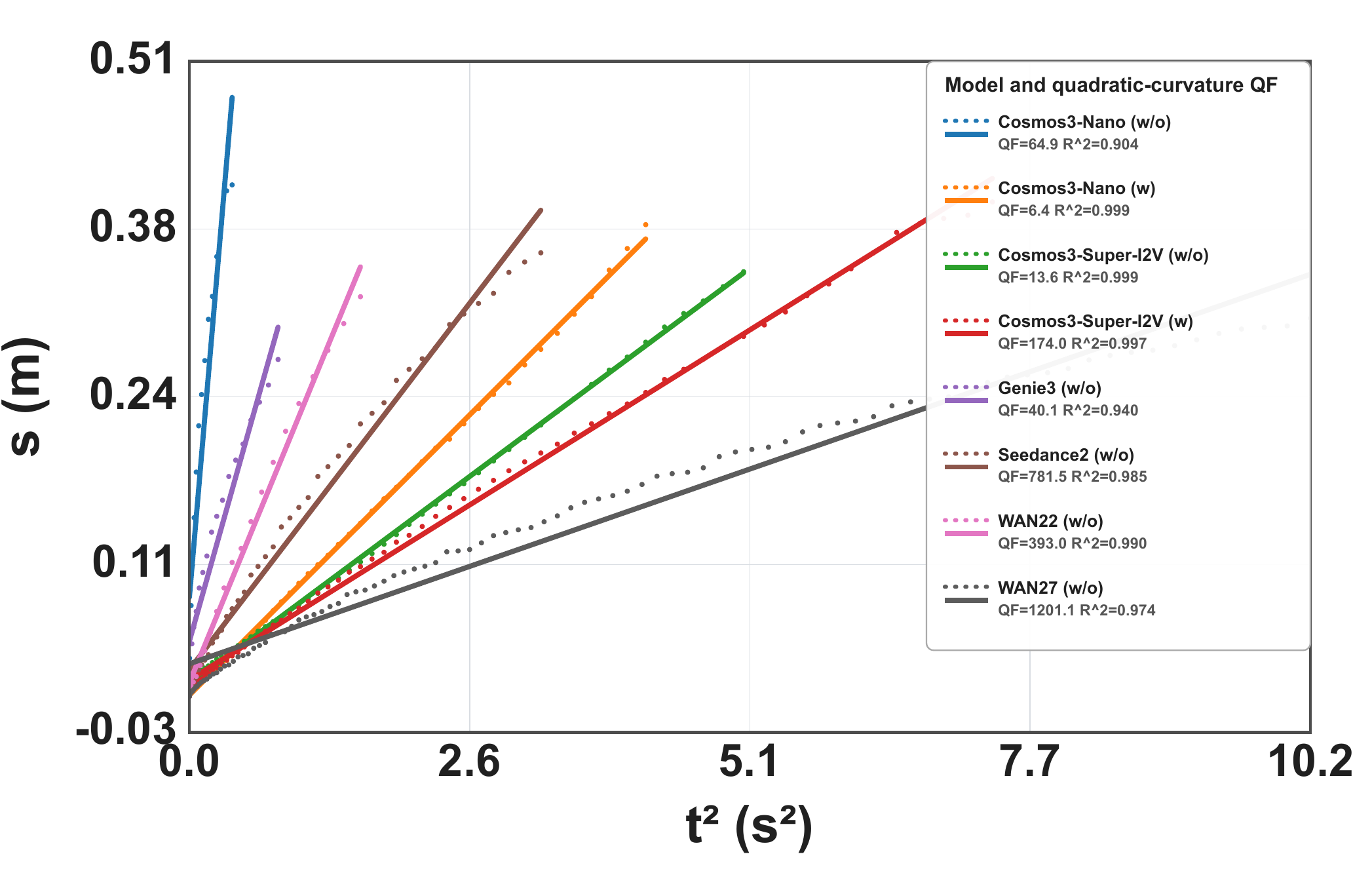}\\
  \small (a) Wood
\end{minipage}
\hfill
\begin{minipage}[t]{0.32\textwidth}
  \centering
  \includegraphics[width=\linewidth]{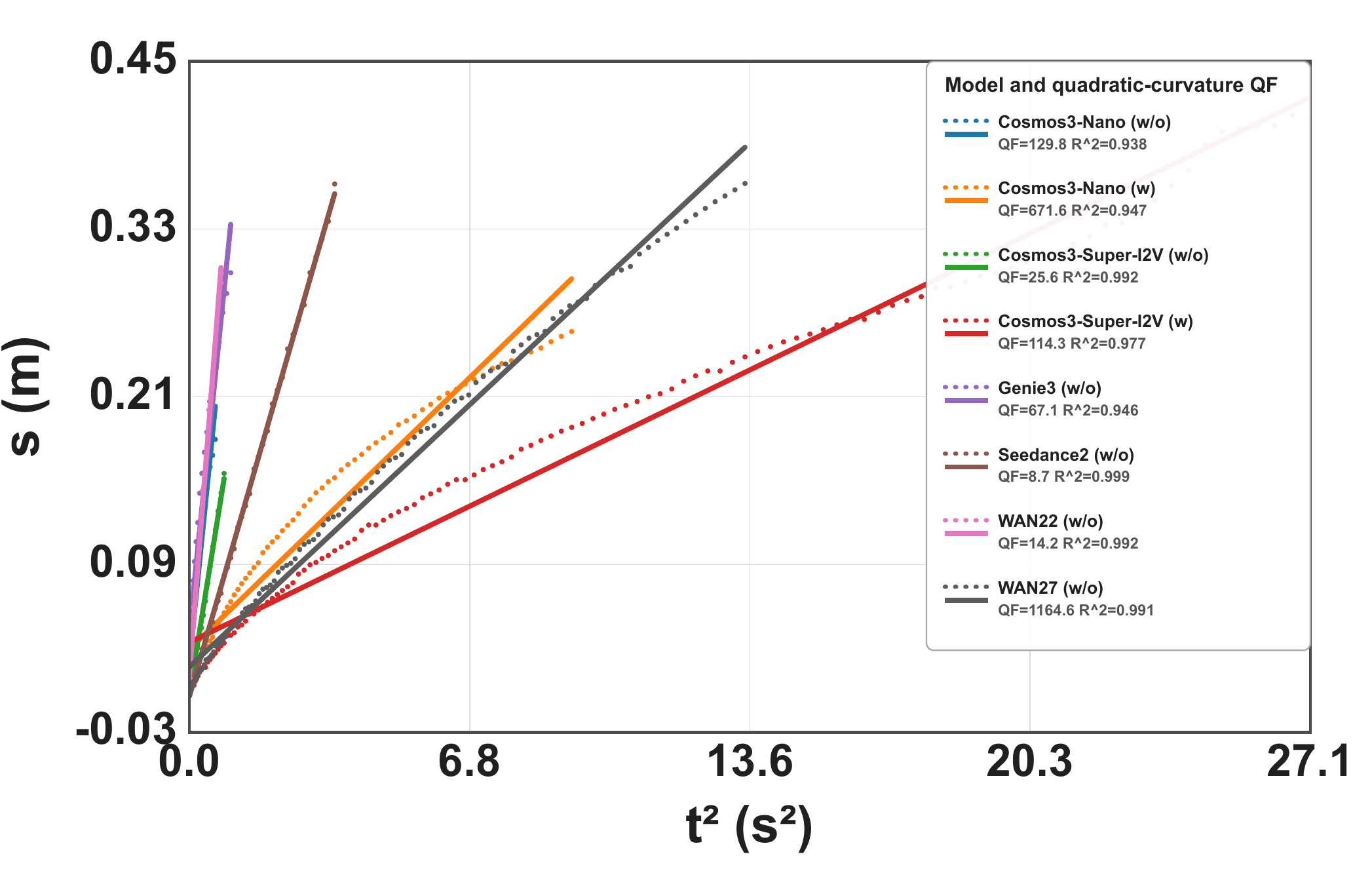}\\
  \small (b) Plastic
\end{minipage}
\hfill
\begin{minipage}[t]{0.32\textwidth}
  \centering
  \includegraphics[width=\linewidth]{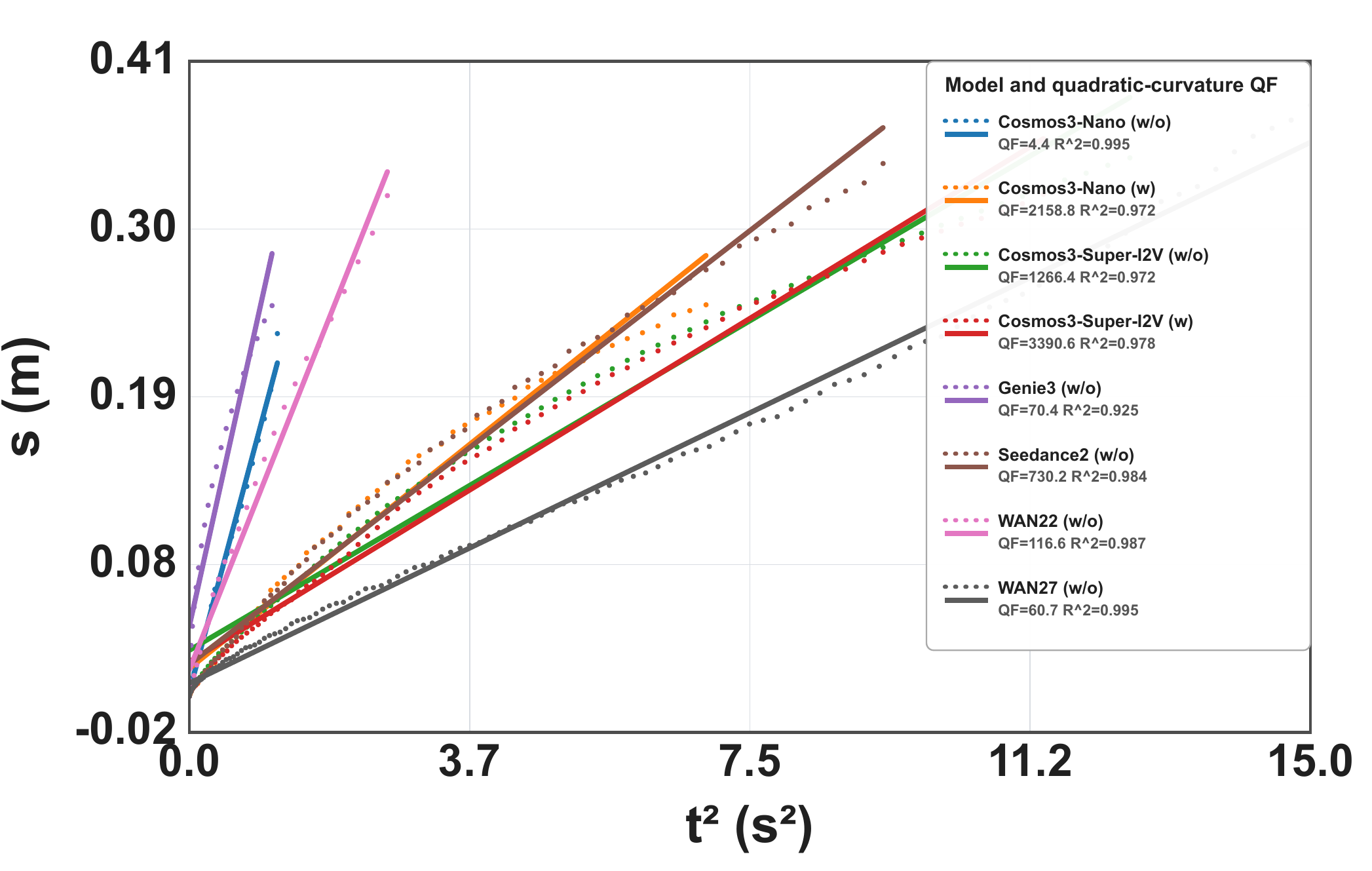}\\
  \small (c) Metal
\end{minipage}
\caption{Slope-slider trajectories for the three materials.  Points are the retained, unfiltered projected displacements, and solid lines are the linear fits in \(s\) versus \(t^2\).  A visually linear trajectory may still have the wrong slope and hence the wrong acceleration.}
\label{s6}
\end{figure*}

\paragraph{Result Analysis}
\cref{s6} shows substantial variation in both slope and curvature across models and materials.  
For wood, Cosmos3-Super-I2V without the negative prompt gives the lowest tabulated QFI of \(13.61\), whereas the closest acceleration is \(2.06\,\mathrm{m/s^2}\) from Cosmos3-Nano without the negative prompt, compared with the real value of \(2.58\,\mathrm{m/s^2}\). 
The model with the best equation form is therefore not the model with the best physical parameter.

The discrepancy is larger for plastic and metal.  
Seedance 2.0 gives the lowest plastic QFI of \(8.69\), but the closest plastic acceleration among all models is only \(0.75\,\mathrm{m/s^2}\), far below the \(2.57\,\mathrm{m/s^2}\) baseline.  
For metal, Cosmos3-Nano without the negative prompt achieves the lowest QFI of \(4.41\), while the closest acceleration is \(0.43\,\mathrm{m/s^2}\) from Genie 3, compared with \(2.67\,\mathrm{m/s^2}\) in the real experiment.  
The different slopes in \cref{s6} visualize this parameter mismatch directly.
Furthermore, the negative prompt has no consistent effect: for example, the wood QFI of Cosmos3-Super-I2V changes from \(13.61\) to \(569.36\), despite improving other model--task pairs.

\subsubsection{Turntable}
\label{app:c4}
\paragraph{Direct Acceleration Analysis}
This task tests whether the horizontal force required by a generated block trajectory is compatible with the available friction on the turntable.  
The known \(0.05\,\mathrm m\) block edge provides the metric scale.  
After removing
only sustained leading stationary frames and terminal frozen frames, the metric trajectory is differentiated directly.  For equally spaced samples with \(\Delta t=1/\mathrm{fps}\),
\begin{equation}
    \mathbf a_i
    =
    \frac{\mathbf p^{\mathrm m}_{i+1}
    -2\mathbf p^{\mathrm m}_{i}
    +\mathbf p^{\mathrm m}_{i-1}}
    {\Delta t^2},
    \qquad
    a_i=\|\mathbf a_i\|_2.
    \label{eq:c4-acceleration}
\end{equation}
For unequal time gaps caused by missing frames, the implemented three-point form is
\begin{equation}
    \mathbf a_i
    =
    \frac{2}{t_{i+1}-t_{i-1}}
    \left[
    \frac{\mathbf p^{\mathrm m}_{i+1}-\mathbf p^{\mathrm m}_{i}}
    {t_{i+1}-t_i}
    -
    \frac{\mathbf p^{\mathrm m}_{i}-\mathbf p^{\mathrm m}_{i-1}}
    {t_i-t_{i-1}}
    \right].
    \label{eq:c4-unequal}
\end{equation}
This local calculation is used because a single fixed acceleration-vector fit would allow the continuously changing centripetal direction to cancel over a rotation.  
No position smoothing is applied before differentiation.

\paragraph{Dynamic Error}
For block mass \(m\), friction coefficient \(\mu\), and \(g=9.80665\,\mathrm{m/s^2}\), the maximum available friction magnitude is
\begin{equation}
    F_{\mathrm{fric,max}}=\mu mg.
    \label{eq:c4-friction}
\end{equation}
The positive-part operator is denoted by \([z]_+=\max(0,z)\).  The per-frame force exceedance and the Dynamic Error (DE) reported in the main table are
\begin{equation}
    e_i=\left[m a_i-F_{\mathrm{fric,max}}\right]_+,
    \qquad
    \mathrm{DE}=\frac{1}{M}\sum_{i=1}^{M}e_i.
    \label{eq:c4-de}
\end{equation}
Thus DE has units of newtons and is zero whenever the required force does not exceed the friction limit.  
The implementation also records the median, 95th-percentile, and maximum accelerations, the fraction of frames exceeding the limit, and the maximum force exceedance.  
The main-table entries use the mean in \cref{eq:c4-de}; for example, this is why small isolated acceleration spikes affect DE without defining it solely by the maximum spike.

As an auxiliary geometric diagnostic, the radial range about the calibrated turntable center is divided by the block edge length.  
A range above \(10\%\) flags possible radial sliding.  Pure tangential slip cannot be identified without an independently tracked turntable angle, so the radial flag is not substituted for DE.

\paragraph{Result Analysis}
The turntable task produces several zero or near-zero DE values.  
This means that the frame-wise force estimated from the generated motion generally does not exceed the calibrated friction limit; it does not prove that the block co-rotates correctly with the turntable.  
For example, the wood results include exact zero values for several configurations, while Cosmos3-Nano without the negative prompt has \(\mathrm{DE}=0.078\,\mathrm N\) and Wan-2.7 without the negative prompt reaches \(0.18\,\mathrm N\).  
For metal, Wan-2.7 with the negative prompt gives the largest reported value, \(0.78\,\mathrm N\), whereas several other configurations remain at or near zero.

These results indicate that abrupt centroid changes and excessive local acceleration are detected by DE, but overly slow, nearly stationary, or incorrectly synchronized motion can still obtain \(\mathrm{DE}=0\).  
The radial-sliding flag and direct trajectory inspection are therefore necessary complements.  
The paired values also show that the negative prompt is not uniformly beneficial: it may suppress acceleration exceedance for one model and material while increasing it for another.

\subsubsection{Bouncing Ball}
\label{app:c19}
\paragraph{First Free-Fall Segment}
This metric evaluates only the first airborne descent before collision or bounce.  
The common reference frame contains a ball of known diameter \(D_{\mathrm{ball}}=0.06\,\mathrm m\).  
If its mean annotated pixel diameter is \(\overline d_{\mathrm{px}}\), the scale is
\begin{equation}
    c=\frac{D_{\mathrm{ball}}}{\overline d_{\mathrm{px}}}.
    \label{eq:c19-scale}
\end{equation}
Image-down is defined as the positive vertical direction.  
Relative to the retained release point,
\begin{equation}
    s_i=c(\widetilde y_i-\widetilde y_0),
    \qquad
    q_i=c(\widetilde x_i-\widetilde x_0),
    \label{eq:c19-displacement}
\end{equation}
where \(s_i\) is the vertical displacement used by the main metric and \(q_i\) is retained as a horizontal-drift diagnostic.

A sustained-window detector locates release after any initial wait.  
Once downward motion is established, the first vertical reversal, sustained contact plateau, tracking transfer, or loss of observations ends the ballistic segment.  
Normal post-impact samples are excluded rather than penalized,
because a bounce does not obey the single free-fall model.  
All samples within the selected airborne interval remain unfiltered.

\paragraph{Acceleration and QFI}
Static release under constant gravity predicts
\begin{equation}
    s(t)=s_0+\frac{1}{2}at^2,
    \qquad
    a=g
    \label{eq:c19-freefall}
\end{equation}
in the ideal case.  
The same constrained linear fit as in \cref{eq:c3-acceleration} is applied:
\begin{equation}
    \widehat{s}_i=b+kt_i^2,
    \qquad
    a=2k.
    \label{eq:c19-acceleration}
\end{equation}
The reported acceleration is compared with the real baseline \(g\approx9.81\,\mathrm{m/s^2}\).  
A visually smooth parabola can therefore have low shape error but still encode an incorrect gravity magnitude.

For consistency with Slope slider, QFI is obtained by adding a \(t^4\) term and using exactly \cref{eq:qfi}.  
Here \(\mathrm{SSE}_{L}\) is the residual sum of squares of \cref{eq:c19-acceleration}, and \(\mathrm{SSE}_{Q}\) is that of \(\widehat s=b_Q+k_Qt^2+c_Qt^4\).  
Low QFI indicates that the expected linear \(s\)-versus-\(t^2\) form is sufficient; it does not by itself imply that the fitted slope yields the correct gravitational acceleration.

The implementation additionally diagnoses horizontal drift by fitting
\begin{equation}
    q(t)=b_x+v_xt+\frac{1}{2}a_xt^2
    \label{eq:c19-horizontal}
\end{equation}
and recording horizontal range, net displacement, and \(a_x/g\).  These diagnostics prevent an accurate-looking vertical component from hiding a strongly non-vertical generated trajectory.

\paragraph{Result Analysis}

\cref{fig4}(b) contains both nearly linear segments and conspicuous curvature, plateaus, reversals, and isolated jumps.  
Cosmos3-Super-I2V with the negative prompt obtains the lowest tabulated QFI of \(12.50\), improving substantially over its value of \(270.69\) without the negative prompt.
Nevertheless, its fitted acceleration is only \(0.088\,\mathrm{m/s^2}\), which is less than one percent of the
\(9.81\,\mathrm{m/s^2}\) real baseline.  
Thus, its trajectory is closer to a straight line in \(s\) versus \(t^2\), but the line has the wrong slope.

Seedance 2.0 provides the closest acceleration, \(1.84\,\mathrm{m/s^2}\), but this remains about \(81\%\) below gravity and its QFI is \(65.16\).  
Genie 3 produces a negative fitted acceleration of
\(-0.052\,\mathrm{m/s^2}\), consistent with the reversed trend visible in the curve.  
Wan-2.2 exhibits a long shallow region followed by stronger motion, which leads to a large QFI of \(428.39\).  
Overall, no evaluated model simultaneously recovers both the expected free-fall form and the correct gravitational scale.

\subsubsection{Newton's Cradle}
\label{app:c5}
\paragraph{Apex-Based Momentum Transfer}
For equal masses, the ratio of outgoing to incoming momentum can be estimated from the gravitational potential heights at the corresponding pendulum apices.  
If the incoming left ball is released from height \(h_L\) above the resting ball-center level, conservation of mechanical energy gives its pre-impact speed \(v_L=\sqrt{2gh_L}\).  For the \(i\)-th detected post-impact apex of the right ball at height \(h_{R,i}\), the corresponding outgoing speed is \(v_{R,i}=\sqrt{2gh_{R,i}}\).  
The event-level Momentum Transfer Efficiency (MTE) is therefore
\begin{equation}
    \eta_i
    =\frac{m v_{R,i}}{m v_L}
    =\sqrt{\frac{h_{R,i}}{h_L}},
    \label{eq:c5-event-mte}
    \end{equation}
    and the main-table value averages all valid right-ball apex events:
    \begin{equation}
    \mathrm{MTE}
    =\overline{\eta}
    =\frac{1}{N}\sum_{i=1}^{N}\eta_i.
    \label{eq:c5-mte}
\end{equation}

The reference ball diameter is \(0.02\,\mathrm m\), which is used separately for each generated resolution to convert vertical pixel differences into heights.  
The annotated initial left-ball height is approximately \(h_L=13.294\,\mathrm{cm}\).  
Right-ball events are local high/right apices identified after independent per-frame detection, with a minimum temporal separation to avoid counting multiple neighboring frames as separate collisions.  
Summary images are not counted as events.

An ideal lossless equal-mass cradle has \(\mathrm{MTE}\approx1\). 
Values below one indicate incomplete momentum transfer or excessive damping, while a value above one indicates generated amplitude gain.  
A dash in the main table means that the model did not yield a valid sequence of independently identifiable right-ball apex events, rather than an MTE of zero.

\paragraph{Result Analysis}
Six of the ten model--prompt configurations do not produce a valid sequence from which MTE can be measured.  
Among the valid generations, Wan-2.2 with the negative prompt gives the highest mean efficiency, \(\mathrm{MTE}=0.76\).  
Wan-2.7 without and with the negative prompt reaches \(0.55\) and \(0.48\), respectively, while Seedance 2.0 reaches only \(0.24\).
All valid averages are below the ideal value of one, indicating incomplete transfer to the outgoing ball.
Although the first detected Wan-2.2 event slightly exceeds one, its later events decay strongly, reducing the event-averaged efficiency.  
The high failure rate and large inter-event variation show that generating a visually recognizable cradle is not sufficient to reproduce repeated collision dynamics.

\subsubsection{Pendulum}
\label{app:c7}
\paragraph{Angle Extraction and Geometric Constraint}
This task uses a calibrated fixed pivot \(\mathbf p=(x_p,y_p)\), a ball diameter of \(0.06\,\mathrm m\), and the reference pivot-to-ball-center length.  
After resolution normalization, the signed angle relative to image-vertical-down is
\begin{equation}
    \theta_i
    =
    \operatorname{atan2}
    \left(\widetilde{x}_i-x_p,\widetilde{y}_i-y_p\right).
    \label{eq:c7-angle}
\end{equation}
The instantaneous radius
\begin{equation}
    r_i=\sqrt{(\widetilde{x}_i-x_p)^2+(\widetilde{y}_i-y_p)^2}
    \label{eq:c7-radius}
\end{equation}
is also retained to diagnose pivot drift and changing pendulum length.  
Only a sustained leading release wait is removed.  Turning points, damping, amplitude growth, temporary stalls, and other behavior after release remain in the fit.

\paragraph{Damped-Oscillation Fit}
The small-angle damped-pendulum equation is
\begin{equation}
    \ddot{\theta}+2\beta\dot{\theta}+\omega_0^2\theta=0,
    \qquad
    \omega_0=\sqrt{\frac{g}{L}}.
    \label{eq:c7-dynamics}
\end{equation}
For an underdamped trajectory, the fitted model is
\begin{equation}
    \widehat{\theta}(t)
    =\theta_{\mathrm{eq}}
    +e^{-\beta t}
    \left[
    C\cos(\omega_dt)+D\sin(\omega_dt)
    \right].
    \label{eq:c7-fit}
\end{equation}
For each candidate pair \((\beta,\omega_d)\), the coefficients \(\theta_{\mathrm{eq}},C,D\) are solved by linear least squares; the search then selects the pair with the smallest residual sum of squares.  
The fit does not force \(\beta\) to be positive, so nonphysical amplitude growth is not hidden.

The coefficient of determination reported in the main table is
\begin{equation}
    R^2
    =1-
    \frac{\sum_{i=1}^{N}
    (\theta_i-\widehat{\theta}_i)^2}
    {\sum_{i=1}^{N}
    (\theta_i-\overline{\theta})^2}.
    \label{eq:c7-r2}
\end{equation}
High \(R^2\) means that a damped sinusoid explains the generated angular variation, but it does not establish that the period, pivot, or pendulum length is physically correct.

\paragraph{Period Duration}
The Period Duration (PD) is obtained directly from the fitted damped angular frequency:
\begin{equation}
    \mathrm{PD}=T_d=\frac{2\pi}{\omega_d}.
    \label{eq:c7-pd}
\end{equation}
It is compared with the measured real-world baseline of \(1.06\,\mathrm s\) in the main table.  
As a diagnostic, same-side turning points also provide cycle estimates \(T_k=t_{k+2}-t_k\), but the tabulated PD is the global fitted value in \cref{eq:c7-pd}.  Consequently, a sequence can obtain a high \(R^2\) by looking periodic while still receiving an incorrect PD because its oscillation time scale is too slow or too fast.

The fitted frequency and damping also imply
\begin{equation}
    \omega_0=\sqrt{\omega_d^2+\beta^2},
    \qquad
    g_{\mathrm{fit}}=L\omega_0^2,
    \label{eq:c7-effective-gravity}
\end{equation}
which is retained as an auxiliary absolute-scale diagnostic.  
Fixed-pivot radial error and period stability are likewise recorded separately; neither is absorbed into \(R^2\).

\paragraph{Result Analysis}
The trajectories in \cref{fig4}(c) differ substantially in amplitude, equilibrium offset, damping behavior, and period.  Wan-2.2 with the negative prompt and Genie 3 both achieve \(R^2=0.99\), showing that a damped sinusoid describes their angular variation well.  
Their fitted periods are \(1.93\,\mathrm s\) and \(1.90\,\mathrm s\), however, compared with the \(1.06\,\mathrm s\) real baseline.  
Their high shape scores therefore do not imply a correct physical time scale.

The closest period is \(1.83\,\mathrm s\), obtained by Wan-2.7 without the negative prompt, but it is still approximately \(73\%\) longer than the real period and its \(R^2\) is only \(0.71\).  
At the other extreme, Cosmos3-Super-I2V without the negative prompt has a fitted period of \(17.95\,\mathrm s\) and \(R^2=0.50\), reflecting its nearly stationary
angular trajectory.  
The negative prompt sometimes improves the oscillatory
fit or period but does not consistently recover the real dynamics.  
These results support reporting \(R^2\) and PD separately.

\stopcontents[appendix]
\end{document}